\documentclass[10pt,twocolumn,letterpaper]{article}

\usepackage{cvpr}      

\usepackage{graphicx}
\usepackage{amsmath}
\usepackage{amssymb}
\usepackage{booktabs}
\usepackage[numbers,sort,compress]{natbib}

\usepackage[utf8]{inputenc} 
\usepackage[T1]{fontenc}    
\usepackage{url}            
\usepackage{booktabs}       
\usepackage{amsfonts}       
\usepackage{nicefrac}       
\usepackage{microtype}      
\usepackage{overpic}        
\usepackage{algorithm}
\usepackage{algpseudocode}
\usepackage{listings}
\usepackage{wrapfig}
\usepackage{amsthm}
\usepackage{colortbl}
\usepackage{makecell}
\usepackage{multirow}
\usepackage{tabularx}
\usepackage{verbatim}
\usepackage[dvipsnames]{xcolor}
\usepackage[pagebackref=true,breaklinks=true,colorlinks,bookmarks=false,citecolor=blue,linkcolor=blue]{hyperref}

\usepackage[accsupp]{axessibility} 

\graphicspath{{./images/}}
\DeclareGraphicsExtensions{.pdf,.jpg,.png}

\usepackage[capitalize]{cleveref}
\crefname{equation}{Eq.}{Eq.}
\crefname{figure}{Fig.}{Fig.}
\crefname{table}{Tab.}{Tab.~}
\crefname{section}{Sec.}{Sec.~}
\crefname{algorithm}{Alg.}{Alg.~}
\crefname{thm}{Theorem}{Theorem~}
\crefname{lemma}{Lemma}{Lemma~}
\crefname{appendix}{Appendix}{Appendix~}
\newtheorem{theorem}{Theorem}

\def\ie{\textit{i.e.,~}}
\def\eg{\textit{e.g.,~}}
\def\etc{\textit{etc~}}

\def\sota{state-of-the-art~}

\definecolor{tomato}{rgb}{1.0, 0.39, 0.28}
\definecolor{cornflowerblue}{rgb}{0.39, 0.58, 0.93}

\usepackage{amsmath,amsfonts,bm}

\def\eqref#1{equation~\ref{#1}}

\def\1{\bm{1}}

\def\rvc{{\mathbf{c}}}

\def\rvf{{\mathbf{f}}}

\def\rvh{{\mathbf{h}}}
\def\rvu{{\mathbf{i}}}

\def\rvp{{\mathbf{p}}}
\def\rvq{{\mathbf{q}}}

\def\rvu{{\mathbf{u}}}

\def\rvx{{\mathbf{x}}}

\def\rvz{{\mathbf{z}}}

\def\vx{{\bm{x}}}

\def\vz{{\bm{z}}}

\DeclareMathAlphabet{\mathsfit}{\encodingdefault}{\sfdefault}{m}{sl}
\SetMathAlphabet{\mathsfit}{bold}{\encodingdefault}{\sfdefault}{bx}{n}

\def\gC{{\mathcal{C}}}

\def\gF{{\mathcal{F}}}

\def\gH{{\mathcal{H}}}

\def\gL{{\mathcal{L}}}

\newcommand{\tbf}[1]{\textbf{#1}}
\newcommand{\ul}[1]{\underline{#1}}

\def\f{f_\theta}
\newcommand{\F}[1]{f_\theta \left(#1\right)}
\def\zstar{\rvz^*}

\def\x{\rvx}

\def\p{\rvp}
\def\c{\rvc}
\def\u{\rvu}
\def\q{\rvq}
\def\h{\rvh}
\def\flow{\rvf}
\def\hstar{\rvh^*}
\def\fstar{\rvf^*}

\newcommand{\Rdim}[1]{\mathbb{R}^{#1}}

\def\ExactGrad{\frac{\partial \gL}{\partial \theta}}
\def\PhantomGrad{\widehat{\frac{\partial \gL}{\partial \theta}}}

\def\J{\left( I - \frac{\partial \f}{\partial \zstar} \right)}

\def\invJ{\J^{-1}}

\newcommand{\emptybox}[2][\textwidth]{%
  \begingroup
  \setlength{\fboxsep}{-\fboxrule}%
  \noindent\framebox[#1]{\rule{0pt}{#2}}%
  \endgroup
}

\definecolor{codeblue}{rgb}{0.25, 0.5, 0.5}
\definecolor{codekw}{rgb}{0.35, 0.35, 0.75}
\lstdefinestyle{Pytorch}{
    language         = Python,
    backgroundcolor  = \color{white},
    basicstyle       = \ttfamily\footnotesize,
    columns          = fullflexible,
    breaklines       = true,
    captionpos       = b,
    commentstyle     = \fontsize{4pt}{4pt}\color{codeblue},
    keywordstyle     = \fontsize{4pt}{4pt}\color{codekw},
    morekeywords     = with,
}

\makeatletter
\apptocmd\@maketitle{{\vspace{-10pt}\mainfigure{}\vspace{10pt}\par}}{}{}
\makeatother

\newcommand\mainfigure{%
\centering
\vspace{-.2cm}
\includegraphics[width=\textwidth]{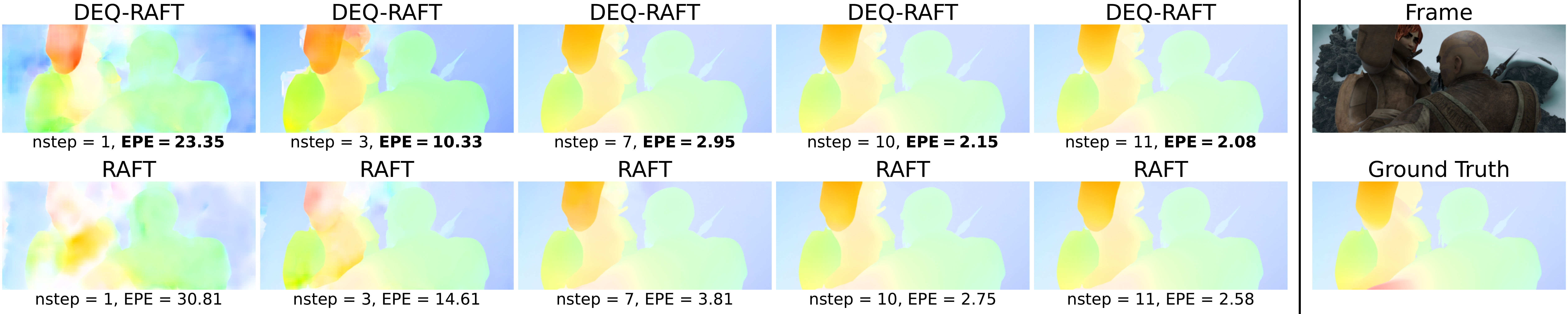}
\vspace{-.6cm}
\captionof{figure}{%
\textbf{A deep equilibrium (DEQ) flow estimator directly models the flow as a path-independent, ``infinite-level'' fixed-point solving process.} We propose to use this implicit framework to replace the existing recurrent approach to optical flow estimation. The DEQ flows converge faster, require less memory, are often more accurate, and are compatible with prior model designs like RAFT~\cite{RAFT}.
}
\label{fig:demo}
}

\def\cvprPaperID{11243} 
\def\confName{CVPR}
\def\confYear{2022}

\begin{document}

\title{Deep Equilibrium Optical Flow Estimation}

\author{
    Shaojie Bai\textsuperscript{1}\footnotemark[1]
    \qquad
    Zhengyang Geng\textsuperscript{2}\footnotemark[1]
    \qquad
    Yash Savani\textsuperscript{1}
    \qquad
    J. Zico Kolter\textsuperscript{1,3}
    \\
    \quad $^1$Carnegie Mellon University \quad $^2$Peking University \quad $^3$Bosch Center for AI
    \\
    {\tt\small \{shaojieb,ysavani,zkolter\}@cs.cmu.edu} \quad {\tt\small zhengyanggeng@gmail.com}
}

\maketitle

\begin{abstract}
    Many recent state-of-the-art (SOTA) optical flow models use finite-step recurrent update operations to emulate traditional algorithms by encouraging iterative refinements toward a stable flow estimation.
    However, these RNNs impose large computation and memory overheads, and are not directly trained to model such ``stable estimation''. They can converge poorly and thereby suffer from performance degradation.
    To combat these drawbacks, we propose deep equilibrium (DEQ) flow estimators, an approach that directly solves for the flow as the infinite-level fixed point of an implicit layer (using any black-box solver)~\cite{DEQ}, and differentiates through this fixed point analytically (thus requiring $O(1)$ training memory).
    This implicit-depth approach is not predicated on any specific model, and thus can be applied to a wide range of SOTA flow estimation model designs (\eg RAFT~\cite{RAFT} and GMA~\cite{GMA}). 
    The use of these DEQ flow estimators allows us to compute the flow faster using, \eg fixed-point reuse and inexact gradients, consumes $4\sim6\times$ less training memory than the recurrent counterpart, and achieves better results with the same computation budget. 
    In addition, we propose a novel, sparse fixed-point correction scheme to stabilize our DEQ flow estimators, which addresses a longstanding challenge for DEQ models in general.
    We test our approach in various realistic settings and show that it improves SOTA methods on Sintel and KITTI datasets with substantially better computational and memory efficiency.
\end{abstract}

\renewcommand{\thefootnote}{\fnsymbol{footnote}}
\footnotetext[1]{Equal contribution. Our \href{https://github.com/locuslab/deq-flow}{code} is available.}
\renewcommand*{\thefootnote}{\arabic{footnote}}

\section{Introduction}
\label{sec:intro}
Optical flow estimation is the classic computer vision task of predicting the pixel-level motions between video frames~\cite{lucas1981iterative,horn1981determining,flownet,maskflownet,RAFT}. Learning-based approaches to this problem, which outperformed classical approaches, proposed the use of conventional deep convolutional networks to learn a flow estimate~\cite{flownet,liteflownet2,maskflownet}.
Recent progress has shown that finite-step, unrolled and recurrent update operations significantly improve the estimation performance, exemplified by the emergence of the RAFT~\cite{RAFT} method. Contemporary optical flow models that employ this approach typically rely on a Gated Recurrent Unit (GRU)~\cite{GRU} to \emph{iteratively refine} the optical flow estimate. 
This approach was motivated to emulate traditional optimization-based methods, and the update operators defined accordingly have become the standard design for state-of-the-art flow models~\cite{RAFT,autoflow,GMA,eldesokey2021normalized,Jiang2021LearningOF}.

Despite their superior performance, these rolled-out recurrent networks suffer from a few drawbacks.
First, training these models involves tracking a long hidden-state history in the backpropagation-through-time (BPTT) algorithm~\cite{werbos1990backpropagation}, which yields a significant computational and memory burden. Therefore, these models tend to scale poorly with larger images and more iterations. 
Second, although these models were designed to emulate \emph{traditional optimization approaches} which solve for a ``stable estimate" with as many steps as needed, the recurrent networks do not directly model such a minimum-energy optima state. Rather, they stop after a predefined $L$ update steps, and are still trained in a path-dependent way using BPTT. We also show later in Fig.~\ref{fig:deq-reuse} that the GRUs frequently oscillate instead of converging.

In this work, we introduce deep equilibrium (DEQ) flow estimators based on recent progresses in implicit deep learning, represented by DEQ models~\cite{DEQ,MDEQ,MON,kawaguchi2020theory,Kolter2020}. 
Our method functions as a superior and natural framework to replace the existing recurrent, unrolling-based flow estimation approach. 

There are multiple reasons why this method is preferable.
\textbf{First}, instead of relying on the na\"ive iterative layer stacking, DEQ models define their outputs as the fixed points of a single layer $\f$ using the input $\x$, \ie $\zstar = \F{\zstar, \x}$, modeling an ``infinite-layer'' equilibrium representation. We can directly solve for the fixed point using specialized black-box solvers, \eg quasi-Newton methods~\cite{broyden1965class,anderson1965}, in a spirit much more consistent with the traditional optimization-based perspective~\cite{fleet2006optical,horn1981determining}. This approach expedites the stable flow estimation process while often yielding better results.
\textbf{Second}, we no longer need to perform BPTT. Instead, DEQ models can directly differentiate through the final fixed point $\zstar$ without having to store intermediary states during the forward computation, considerably lowering the training memory cost.
\textbf{Third}, this fixed-point formulation justifies numerous implicit network enhancements such as 1) fixed-point reuse from adjacent video frames; and 2) inexact gradients~\cite{Ham,SamyFPN,PhantomGrad}. The former helps avoid redundant computations, thus substantially accelerating flow estimations; and the latter makes the backward pass computationally \emph{almost free}!
\textbf{Fourth}, the DEQ approach is not predicated on any specific structure for $\f$. Therefore, DEQ is a \emph{framework} that applies to a wide range of these SOTA flow estimation model designs (\eg RAFT~\cite{RAFT}, GMA~\cite{GMA}, and Depthstillation~\cite{aleotti2021learning}), and we can obtain the aforementioned computational and memory benefits with even additional gain based on the specific structure of $f_\theta$.

In addition to suggesting DEQ flow estimators as a superior replacement to the existing recurrent approach, we also tackle the longstanding instability challenge of training DEQ networks~\cite{chen2018neural,DEQ,DEQ_JR,MON}. Inspired by the RAFT model, we propose a novel, sparse fixed-point correction scheme that substantially stabilizes our DEQ flow estimators.

The contributions of this paper are as follows. 
\textbf{First}, we propose the deep equilibrium (DEQ) approach as a new natural starting point for formulating optical flow methods. A DEQ approach directly models and substantially accelerates the fixed-point convergence of the flow estimation process, avoids redundant computations across video frames, and comes with an almost-free backward pass.
\textbf{Second}, we show that the DEQ approach is orthogonal to, and thus compatible with, the prior modeling efforts (which focus on the model design and feature extraction)~\cite{RAFT,GMA} and data-related efforts~\cite{autoflow}. With DEQ, these prior arts are now more computationally and memory efficient as well as more accurate. For instance, on KITTI-15~\cite{kitti} (\emph{train}) a zero-shot DEQ-based RAFT model further reduces the state-of-the-art F1-all measure by $21.0\%$ while using the underlying RAFT design.
\textbf{Third}, we introduce a sparse fixed-point correction scheme that significantly stabilizes DEQ models on optical flow problems while only adding minimal cost, and show that on flow estimation tasks this approach is superior to the recently proposed Jacobian-based regularization~\cite{DEQ_JR}.

\section{Related Work}
\label{sec:related_work}

\paragraph{Iterative Optical Flow.} Although optical flow is a classic problem, there has recently been substantial progress in the area.
Earlier methods~\cite{horn1981determining,black1993framework,Zach2007ADB,Wedel2008AnIA,Brox2011LargeDO} formulated the optical flow prediction as an energy minimization problems using continuous optimization with different objective terms. This perspective inspired multiple improvements that used discrete optimization to model optical flows, \ie those based on conditional random fields~\cite{Menze2015DiscreteOF}, global optimization~\cite{Chen2016FullFO}, and inference on the global 4D cost volume~\cite{Xu2017AccurateOF}. 
More recently, with the advancement of deep learning, there have been an explosion of efforts trying to emulate these optimization steps via deep neural networks. For example, a number of optical flow methods are based on deep architectures that rely on coarse-to-fine pyramids~\cite{pwcnet,pwcnet+,liteflownet,liteflownet2,flownet,ilg2017flownet,vcn}. Specifically, recent research efforts have turned to iterative refinements, which typically involves stacking multiple direct flow prediction modules~\cite{ilg2017flownet,ranjan2017optical}. The RAFT model~\cite{RAFT}, which inspired this work, first showed they could achieve state-of-the-art performance on optical flow estimation using a global correlation volume and a ConvGRU update operator that mimics the behavior of traditional optimizers, which tends to converge to a stable flow estimate. Built on top of this recurrent unrolling framework of RAFT, \citet{GMA}, and \citet{Zhang2021SepFlow} introduced additional attention-style modules prior to the recurrent stage to improve the modeling of occlusions and textureless areas. Another contemporary work, AutoFlow~\cite{autoflow}, exploits bilevel optimization to automatically render and augment training data for optical flow. Finally, \citet{jiang2021learning} proposes to speed up these flow estimators by replacing the dense correlation volume with a sparse alternative.

The focus of this paper is on a direction that is largely orthogonal to and thus complementary to these modeling efforts. We challenge and improve the ``default'' \emph{recurrent, unrolled} formulation of training flow estimators themselves. With the help of the recent progress in implicit deep learning (see below), we can maintain the same convergent flow estimation formulation while paying substantially less computation and memory costs.

\vspace{-.25cm}
\paragraph{Implicit deep learning.} Recent research has proposed a new class of deep learning architectures that do not have prescribed computation graphs or hierarchical layer stacking like conventional networks. Instead, the output of these implicit networks is typically defined to be the solution of an underlying dynamical system~\citep{Kolter2020,amos2017optnet,chen2018neural,DEQ,elghaoui2019implicit,Ham}. For example, Neural ODEs~\citep{chen2018neural} model infinitesimal steps of a residual block as an ODE flow. A deep equilibrium (DEQ) network~\citep{DEQ} (which primarily inspired this work) is another class of implicit model that directly solves for a fixed-point representation of a shallow layer $f_\theta$ (\eg, a Transformer block) and differentiates through this fixed point without storing intermediate states in the forward pass. This allows one to train implicit networks with \emph{constant} memory, while fully decoupling the forward and backward passes of training. However, it is known that these implicit models suffer from a few serious issues that have been studied by later works, such as computational inefficiency~\citep{chen2018neural,dupont2019augmented}, instability~\citep{chen2018neural,DEQ,DEQ_JR}, and lack of theoretical convergence guarantees~\citep{kawaguchi2020theory,MON}. On a positive note, followup works have also shown that DEQ-based models can achieve competitive results on challenging tasks such as language modeling~\citep{DEQ}, generative modeling~\citep{lu2021implicit}, semantic segmentation~\citep{MDEQ}, etc. However, to the best of our knowledge, these implicit models have not been applied to the task of optical flow estimation. In this paper, we show that this task could substantially benefit from the DEQ formulation as well.

\section{Method}
\label{sec:method}

We start by introducing some preliminaries of existing flow estimators. These modules are typically applied directly on raw image pairs, with the extracted representations then passed into the iterative refinement stage. We use RAFT~\cite{RAFT} as the illustrative example here while noting that cutting-edge flow estimators generally share similar structure (\ie, for context extraction and correlation computations).

\subsection{Preliminaries}
\label{subsec:preliminaries}

Given an RGB image pair $\p^1, \p^2 \in \Rdim{3 \times H \times W}$, an optical flow estimator aims to learn a correspondence $\flow \in \Rdim{2 \times H \times W}$ between two coordinate grids $\c^1, \c^2$ (\ie $\flow = \c^2 - \c^1$), which describes the per-pixel motion between consecutive frames in the horizontal ($dx$) and vertical ($dy$) directions. To process the matched image pair, we first encode features $\u^1, \u^2 \in \Rdim{C \times H \times W}$ of $\p^1, \p^2$, and produce a context embedding $\q$ from the first image $\p^1$. Then, we construct a group of pyramid global correlation tensors $\gC = \left\{\gC^0, \cdots, \gC^{p-1}\right\}$, where $\gC^k \in \Rdim{H \times W \times H / 2^k \times W / 2^k}$ is found by first calculating the inner product between all pairs of hyperpixels in $\u^1$ and $\u^2$ as $\gC^0$, \ie
\begin{equation}
    \gC^0_{ijmn} = \sum_d \u^1_{ijd} \u^2_{mnd}
\end{equation}
followed by downsampling the last two dimensions to produce $\gC^k$ ($k > 0$). The correlation pyramid $\gC$ and context embedding $\q$, which allow the model to infer large motions and displacements in a global sense, are then passed as inputs into the \emph{iterative refinement} stage. 

\begin{figure*}[t]
\centering
\vspace{-.1cm}
\includegraphics[width=1.02\textwidth]{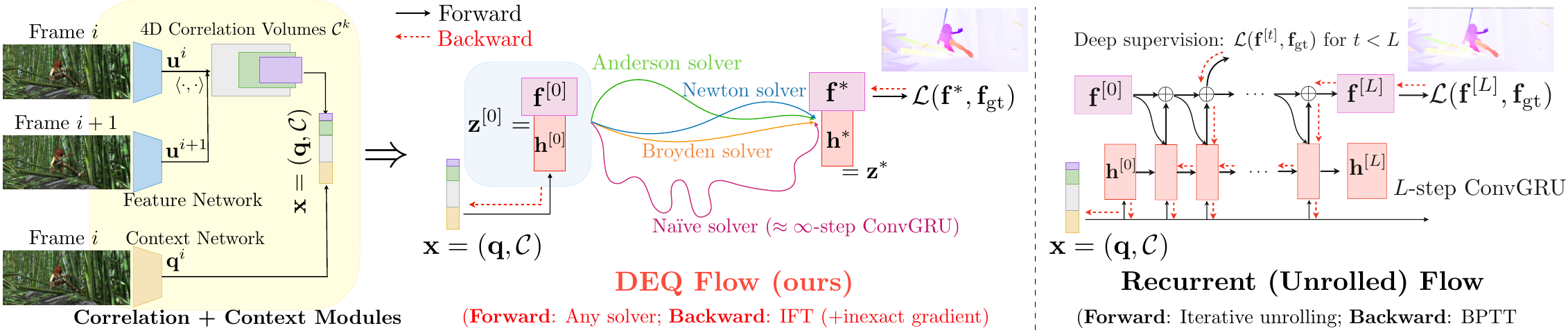}
\caption{A visual comparison of the DEQ flow estimator and the recurrent unrolled flow estimator. After the correlation and context modules (see Sec.~\ref{subsec:preliminaries}), a DEQ flow uses a fast, black-box fixed-point solver (e.g., Anderson) to directly solve for a stable (fixed-point) flow $\zstar=(\hstar, \fstar)$, and differentiate through $\mathbf{z}^\star$ with a cheap inexact gradient. This makes a DEQ flow's backward pass almost free. In contrast, a recurrent flow estimator has to be unrolled for many steps, and needs to perform BPTT, which is costly in both computation and memory.}
\vspace{-.4cm}
\label{fig:recurrent-vs-deq}
\end{figure*}

In this work, we keep the correlation and context computation part intact (see Fig.~\ref{fig:recurrent-vs-deq}) and concentrate on the iterative refinement stage. We refer interested readers to \citet{RAFT} for a more detailed description of the feature extraction process. 

\subsection{Deep Equilibrium Flow Estimator}
\label{subsec:deq-flow-estimator}

Due to the inherent challenges of the flow estimation task, prior works have shown that explicit neural networks struggle to predict the flow accurately, requiring a prohibitively large number of training iterations~\cite{flownet}. Recent works~\cite{RAFT,GMA,aleotti2021learning} have resorted to mimicking the flavor of traditional optimization-based algorithms~\cite{horn1981determining} with RNNs (\eg convGRUs). However, these methods are still quite different from the traditional methods in a few ways. For example, optimization-based methods 1) have an adaptive and well-defined stopping criteria (\eg whenever they reach the optima); 2) are agnostic to the choice of solver (\eg first- or second-order methods);  and 3) are essentially path-independent (\ie the output alone is the only thing we should need). None of these properties are directly characterized by the finite-step unrolling of recurrent networks.

We propose to close this gap with a DEQ-based approach. Specifically, given the context embedding $\q$ and the pyramid correlation tensor $\gC$, a DEQ flow estimator simultaneously solves for the fixed-point convergence of two alternate streams: 1) a latent representation $\h$, which constructs the flow updates; and 2) the flow estimate $\flow$ itself, whose updates are generically related as follows:
\begin{equation}
\label{eq:deq-flow-abs}
\begin{array}{llll}
& \h^{[t+1]}    & = & \gH (\h^{[t]} \quad, \flow^{[t]}, \q, \gC)  \\
& \flow^{[t+1]} & = & \gF (\h^{[t+1]}    , \flow^{[t]}, \q, \gC). \\
\end{array} 
\end{equation}
This formulation captures the form of prominent flow estimator model designs like RAFT~\cite{RAFT} or GMA~\cite{GMA}. Formally, the input $\x=(\q, \gC)$ and model parameters $f_\theta=(\gH, \gF)$ jointly define a dynamical system that the DEQ flow model can \emph{directly} solve the fixed point for using the following flow update equation in its forward pass:
\begin{equation}
\label{eq:deq}
    (\hstar, \fstar) = \zstar = \F{\zstar, \rvx} = \F{(\hstar, \fstar), \rvx}.
\end{equation}
Intuitively, this corresponds to an ``infinite-depth'' feature representation $\zstar$ where, if we perform one more flow update step $\f$, both flow estimation $\flow$ and latent state $\h$ will not change (thus reaching a fixed point, \ie an ``equilibrium''). Importantly, we can leverage much more advanced root solving methods like quasi-Newton methods (\eg Broyden's method~\cite{broyden1965class} or Anderson mixing~\cite{anderson1965}) to find the fixed point. These methods guarantee a much faster (superlinear) and better-quality convergence than if we perform infinitely many na\"ive unrolling steps (as do recurrent networks but only up to a finite number of steps due to computation and memory constraints). Moreover, we note that prior works on implicit networks have shown that the exact structure of $f_\theta$ subsumes a wide variety of model designs, such as a Transformer block~\cite{DEQ,vaswani2017attention}, a residual block~\cite{MDEQ,he2016deep}, or a graph layer~\cite{gu2020implicit,park2021convergent,liu2021eignn}. Similarly, for the deep equilibrium flow estimator, \cref{eq:deq-flow-abs} engulfs exactly the designs of state-of-the-art optical flow models, which we follow and use without modification. For example, for RAFT~\citep{RAFT},
\begin{equation}
\label{eq:deq-flow-raft}
\begin{array}{llll}
& \x          & = & \text{Conv2d}\left([\q,\, \fstar, \gC(\fstar+\c^0)]\right) \\[0.3mm]
& \hstar      & = & \text{ConvGRU}\left(\hstar, [\x,\, \q] \right)   \\[0.15mm]
& \fstar      & = & \fstar + \text{Conv2d}\left(\hstar \right),        \\
\end{array} 
\end{equation}
where $\gC(\fstar+\c^0)$ stands for the correlation lookup as in RAFT~\cite{RAFT}. We also show in Appendix that GMA~\citep{GMA} can be easily written in a similar update form.

The key question is, how do we update and train a DEQ flow estimator. It turns out that we can directly differentiate through this ``infinite-level'' flow state, $(\hstar, \fstar)$, without any knowledge of the forward fixed-point trajectory:
\begin{theorem}
\label{th:ift}
(Implicit Function Theorem (IFT)~\cite{krantz2012implicit,DEQ}) Given the fixed-point flow representation $\zstar=(\hstar, \fstar)$, the corresponding flow loss $\mathcal{L}(\hstar, \fstar, \mathbf{f}_\text{gt})$ and input $\mathbf{x}=(\q, \gC)$, the gradient of DEQ flow is given by
\begin{align}
    \ExactGrad = \frac{\partial \gL}{\,\partial \zstar} \invJ \frac{\partial \f(\zstar, \mathbf{x})}{\partial \theta}
\end{align}
\end{theorem}
\noindent For the proof, see \citet{DEQ}. Importantly, this theorem enables us to \emph{decouple} the forward and backward passes of a DEQ flow estimator; \ie to perform gradient update, we only need the final output $\zstar$ and do not need to run backpropagation-through-time (BPTT). It means a huge memory reduction: whereas an $L$-step recurrent flow estimator takes $O(L)$ memory to perform BPTT, a DEQ estimator reduces the overhead by a factor of $L$ to be $O(1)$ (\eg RAFT uses $L=12$ for training, so using a DEQ flow can theoretically reduce the iterative refinement memory cost by $12\times$).

To summarize, a DEQ flow's forward pass directly solves a fixed-point flow-update equation via black-box solvers; and its backward pass relies only on the final optimum $\zstar$, which make this flow estimation process much more akin to the optimization-based perspective~\citep{horn1981determining}.

\begin{figure*}[t]
\centering
  \begin{subfigure}[b]{0.64\textwidth}
  \includegraphics[width=1.02\textwidth]{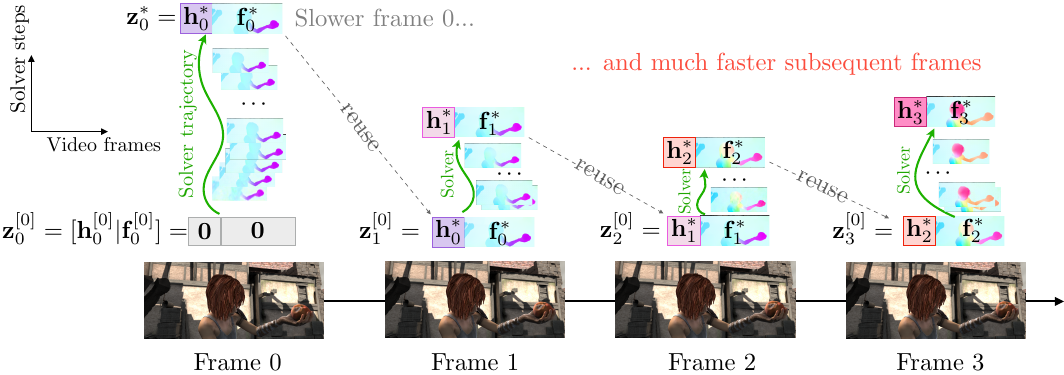}
  \end{subfigure}
  ~
  \begin{subfigure}[b]{0.34\textwidth}
  \includegraphics[width=\textwidth]{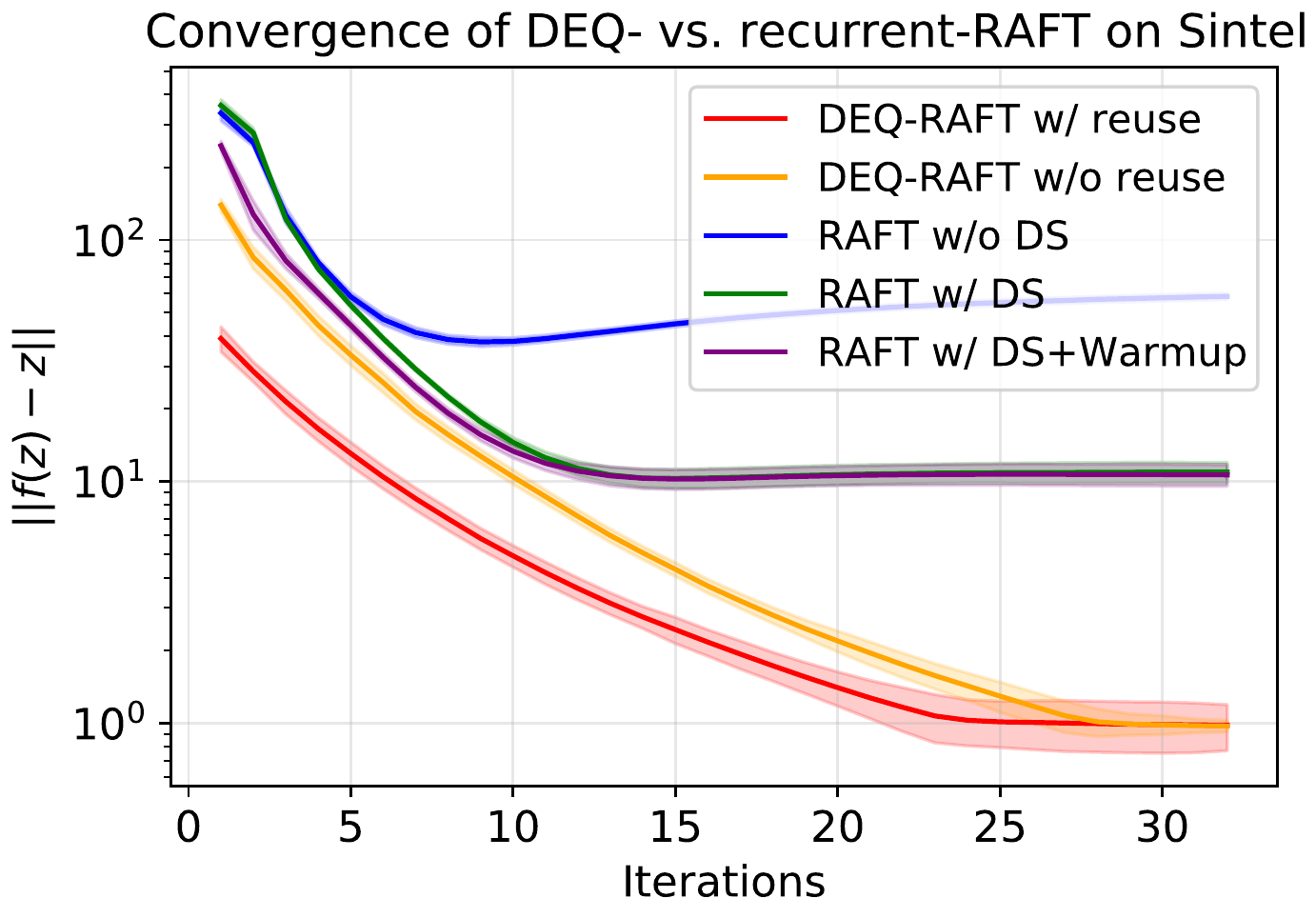}
  \end{subfigure}
\vspace{-.2cm}
\caption{(Left) By reusing fixed-point $\zstar$ from the previous frame's flow estimation, we can ``jump start'' the subsequent equilibrium solving, essentially amortizing the solver cost and speeding up convergence. (Right) Comparing forward convergence of DEQ and recurrent flow estimators on Sintel videos (50 frames). "DS" stands for deep supervision used by RAFT~\cite{RAFT}. DEQ flow with fixed-point reuse converges best; and overall, DEQ flows converge faster than RAFT~\cite{RAFT}.}
\label{fig:deq-reuse}
\vspace{-.4cm}
\end{figure*}

\subsection{Accelerating DEQ Flows}
\label{subsec:accelerate-deq-flow}
Formulating optical flow estimation as a deep equilibrium solution also enables us to fully exploit the toolkit from implicit deep learning. We elaborate below on examples of how this equilibrium formulation can substantially help us improve the forward and backward pipeline and significantly simplify the overall overhead of modern flow estimators.

\vspace{-.5cm}
\paragraph{Inexact Gradients for Training DEQs.} Despite the niceness of the implicit function theorem (IFT), inverting the Jacobian term could quickly become intractable as we deal with high-dimensional feature maps. To combat this, \citet{DEQ} proposed exploiting fast vector-Jacobian products and solving a linear fixed-point system $\mathbf{g}^\top = \mathbf{g}^\top \frac{\partial \f}{\partial \zstar} + \frac{\partial \mathcal{L}}{\partial \zstar}$. However, this approach is still iterative in nature, and in practice, it is no cheaper than the forward flow solving process.

Recent works on implicit networks' backward dynamics~\cite{Ham,SamyFPN,PhantomGrad} suggest that they can typically be trained, and even benefit from, simple approximations of the IFT, while still modeling an ``infinite-depth'' representation through the fixed-point forward pass. That is, we do not need the exact solution to Thm.~\ref{th:ift} to train these networks. Instead we use
\begin{align}
\label{eq:pg}
    \frac{\partial \gL}{\partial \theta} \approx \PhantomGrad = \frac{\partial \gL}{\,\partial \zstar} A \frac{\partial \f(\zstar, \mathbf{x})}{\partial \theta}
\end{align}
where $A$ is a Jacobian (inverse) approximation term. For example,~\cite{Ham,SamyFPN} proposes to use $A=I$ (\ie, 1-step gradient), which simplifies the backward pass of a DEQ flow estimator to $\frac{\partial \gL}{\partial \theta} \approx \frac{\partial \gL}{\,\partial \zstar} \frac{\partial \f(\zstar, \mathbf{x})}{\partial \theta}$. Therefore, unlike the BPTT-based recurrent framework used by existing flow estimators, a DEQ flow estimator's backward pass that uses inexact gradient consists of a single step (and thus is almost free)! Empirically, since we almost eliminate the backward pass cost, the inexact gradients significantly reduce the total training time for DEQ flow estimator further by a factor of almost $2\times$. The capability of using inexact gradients is a direct and unique consequence of the fixed-point formulation and assumes a certain level of stability for the underlying dynamics~\cite{Ham,SamyFPN,PhantomGrad,DEQ_JR}. We discuss an additional approach that further improves the stability of these estimates next.

\vspace{-.35cm}
\paragraph{Sparse fixed-point correction of DEQ flows.} A longstanding challenge in training implicit networks is \emph{the growing instability problem}~\cite{DEQ,DEQ_JR,MON,SamyFPN,PhantomGrad,chen2018neural,dupont2019augmented,kelly2020learning}. In short, since DEQ flow estimators have no discrete layers, they struggle to converge during training.
In other words, the stable flow estimate $\zstar=(\hstar, \fstar)$ could become computationally expensive to reach. This suggests that the optical flow estimation process gets slower during training.

In this work, we propose sparsely applying a fixed-point correction term to stabilize the DEQ flow convergence. Formally, suppose the black-box solver (e.g., Broyden's method) yields a convergence path $(\mathbf{z}^{[0]}, \dots, \mathbf{z}^{[i]}, \dots \zstar)$, where $\mathbf{z}^{[0]}$ is the initial guess and $\zstar$ is the final flow estimate. We then randomly pick $\mathbf{z}^{[i]}=(\mathbf{h}^{[i]}, \mathbf{f}^{[i]})$ on this path (e.g., can be uniformly spaced), and define our total loss to be
\begin{equation}
\label{eq:correction}
\mathcal{L}_\text{total} = \mathcal{L}_\text{main} + \mathcal{L}_\text{cor} = \underbrace{\|\fstar - \mathbf{f}_\text{gt}\|_2^2}_{\text{main loss}} + \gamma \underbrace{\|\mathbf{f}^{[i]} - \mathbf{f}_\text{gt}\|_2^2}_{\text{fixed-point correction}}
\end{equation}
where $\gamma<1$ is a loss weight hyperparameter. This was inspired by the dense step-wise deep supervision used by conventional flow estimators like RAFT~\cite{RAFT}. However, our application here differs in two significant ways. First, we apply this in a sparse manner, with our primary goal being correcting instability. Second, unlike in RAFT, which performs costly BPTT through the RNN chain, this fixed-point correction loss is still \emph{path-independent} and can be understood as a coarse-grained fixed-point estimate. Therefore, we could also perform inexact gradient updates on this correction loss as well; \ie
\begin{equation}
\label{eq:correction-grad}
    \frac{\partial \mathcal{L}_\text{cor}}{\partial \theta} \approx \gamma \frac{\partial \mathcal{L}_\text{cor}}{\partial \mathbf{z}^{[i]}} \frac{\partial \f(\mathbf{z}^{[i]},\x)}{\partial \theta}.
\end{equation}
Empirically, we find this significantly stabilizes the DEQ flow estimator while having no noticeable negative impact on performance. This result is in sharp contrast to existing stabilization methods like Jacobian regularization~\cite{DEQ_JR,hutchinson1989stochastic} which 1) apply only locally to $\zstar$; and 2) usually hurt model performance (see the ablation study in Sec.~\ref{sec:experiments}). Moreover, due to the inexact gradient in Eq.~(\ref{eq:correction-grad}), our method adds almost no extra computation or memory cost. 

While our scope is limited to flow estimation here, we believe this approach suggests a potentially valuable and lightweight solution to the generic instability issue of implicit models, which we leave for future work.

\vspace{-.25cm}
\paragraph{Fixed-point reuse for better initialization.} The DEQ flow estimator's unique formulation also inherits many useful properties from the general optimization framework. One of these nice properties is the ability to perform fixed-point reuse to further accelerate flow estimation convergence. The motivation for this comes from the fact that consecutive frames of a video are typically highly correlated. For instance, perhaps only a few objects are moving in the foreground, while most of the other content and background are nearly identical across these adjacent frames. More formally, if $\mathbf{p}^i$, $\mathbf{p}^{i+1}$, and $\mathbf{p}^{i+2}$ are 3 consecutive video frames, then the ground-truth optical flow $\mathbf{f}_i$ (between $\mathbf{p}^i$ and $\mathbf{p}^{i+1}$) is usually highly correlated to the next ground-truth optical flow $\mathbf{f}_{i+1}$. Thus, when we perform real-time flow estimation with conventional networks like FlowNet~\cite{flownet} and RAFT~\cite{RAFT}, we frequently perform a lot of \emph{redundant} computations. In contrast, with a DEQ flow, we can 
recycle the fixed-point solution $\zstar_i$ of the previous frame, which estimates $\mathbf{f}_i$, as the initial guess $\mathbf{z}_{i+1}^{[0]}$ for the subsequent frame's fixed-point solver. Intuitively, these DEQ flows are able to automatically adjust their forward optimization by exploiting this more informed initial guess, which facilitates convergence speed. It amortizes the cost of flow estimation over long video sequences, since only frame 0 requires full fixed-point solving while the remaining frames can all recycle their predecessor's flow.
We note that such reuse is related to, but still different from the warm-up scheme of RAFT~\cite{RAFT}, which only applies to $\flow$, excludes $\h$, and still has to be unrolled for many steps. In our case, because a DEQ flow directly models a fixed point, such an adaptive computation by exploiting the inductive bias of video data is well-justified.

Fig.~\ref{fig:deq-reuse} shows the practicality of fixed-point reuse on Sintel video sequences. By re-using the fixed point, we can further accelerate the DEQ flow estimator's inference speed by a factor of about $1.6 \times$. Interestingly, while RAFT's iterative unrolling aims to mimic the iterative convergence, we find its activations usually oscillate at a relatively high level after about 15 update iterations.

To summarize, while a conventional recurrent flow estimator like RAFT needs to be unrolled for some finite $L$ steps and back-propagated through the same $L$-step chain, a deep equilibrium flow estimator: 1) leverages the IFT and requires only $O(1)$ training memory, 2) uses inexact gradients to reduce the backward pass to $O(1)$ computation, and 3) can take advantage of correlation between adjacent frames to amortize the flow estimation cost across a long sequence, thus significantly accelerating the forward pass.

\section{Experiments}
\label{sec:experiments}

\setlength\tabcolsep{4pt}
\begin{table*}[t]

\centering
\newcolumntype{C}{>{\centering\arraybackslash}X}
\resizebox{\textwidth}{!}{
\begin{tabularx}{\textwidth}{@{}c l C C C C C C C C@{}}

\toprule
\multirow{2}{*}{Data} & \multirow{2}{*}{Method} & \multicolumn{2}{c}{Sintel (train)} & \multicolumn{2}{c}{KITTI-15 (train)} & \multicolumn{2}{c}{Sintel (test)} & \multicolumn{2}{c}{KITTI-15 (test)}\\
\cmidrule(lr){3-4}
\cmidrule(lr){5-6}
\cmidrule(lr){7-8}
\cmidrule(lr){9-10}
& & Clean & Final & AEPE & F1-all & Clean & Final & F1-fg & F1-all\\

\midrule    
\multirow{14}{*}{C + T} 
    & LiteFlowNet\cite{liteflownet}      & 2.48  & 4.04  & 10.39 & 28.5 & - & - & - & - \\
    & PWC-Net\cite{pwcnet}               & 2.55  & 3.93 & 10.35 & 33.7 & - & - & - & - \\
    & LiteFlowNet2\cite{liteflownet2}    & 2.24  & 3.78  & 8.97 & 25.9 & - & - & - & - \\
    & VCN\cite{vcn}                      & 2.21  & 3.68  & 8.36 & 25.1 & - & -     & - & - \\ 
    & MaskFlowNet\cite{maskflownet}      & 2.25 & 3.61 & - & 23.1 & - & - & - & - \\ 
    & FlowNet2\cite{ilg2017flownet}      & 2.02  & 3.54 & 10.08 & 30.0 & 3.96  & 6.02 & - & - \\
    \cmidrule[\lightrulewidth](r{0.3em}){2-10}
    & RAFT\cite{RAFT}                    & 1.43       & 2.71        & 5.04       & 17.4       & - & - & - & - \\
    & DEQ-RAFT-B                         & 1.48       & 2.81        & 5.01       & 16.3       & - & - & - & - \\
    & DEQ-RAFT-L                         & 1.40       & \ul{2.65}   & 4.76       & 16.1       & - & - & - & - \\
    & DEQ-RAFT-H                         & 1.41       & 2.75        & \ul{4.38}  & \ul{14.9}  & - & - & - & - \\
    & DEQ-RAFT-H$^\dagger$               & 1.34       & \tbf{2.60}  & \tbf{3.99} & \tbf{13.5} & - & - & - & - \\
    \cmidrule[\lightrulewidth](r{0.3em}){2-10}
    & GMA\cite{GMA}                      & \tbf{1.30} & 2.74        & 4.69     & 17.1 & - & - & - & - \\ 
    & DEQ-GMA-B                          & 1.35       & 2.90        & 4.84     & 16.2 & - & - & - & - \\ 
    & DEQ-GMA-L                          & \ul{1.33}  & 2.71        & 4.72     & 16.4 & - & - & - & - \\ 
\midrule
\multirow{6}{*}{C+T+S+K+H}
    & LiteFlowNet2\cite{liteflownet2}    & (1.30) & (1.62) & (1.47) & (4.8)      & 3.48     & 4.69    & 7.62  &  7.62 \\
    & PWC-Net+\cite{pwcnet+}             & (1.71)     & (2.34)  & (1.50) & (5.3) & 3.45     & 4.60    & 7.88  &  7.72 \\
    & VCN \cite{vcn}                     & (1.66)     & (2.24) & (1.16) & (4.1)  & 2.81     & 4.40    & 8.66  &  6.30 \\
    & MaskFlowNet\cite{maskflownet}      & - & - & - & -                         & 2.52     & 4.17    & 7.70  &  6.10 \\
    \cmidrule[\lightrulewidth](r{0.3em}){2-10}
    & RAFT\cite{RAFT}                    & (0.76)  & (1.22)  & (0.63)  & (1.5)   & 1.94       & \tbf{3.18}   & 6.87       & 5.10       \\ 
    & DEQ-RAFT                           & (0.73)  & (1.02)  & (0.61)  & (1.4)   & \tbf{1.82} & 3.23         & \tbf{6.06} & \tbf{4.91} \\
\bottomrule
\end{tabularx}
}
\vspace{-.1cm} 
\caption{\tbf{Evaluation on Sintel and KITTI 2015 datasets.} We report the Average End Point Error~(AEPE), F1-fg (\%), and F1-all (\%) (lower is better). ``C+T'' refers to results that are pre-trained on the Chairs and Things datasets. ``S+K+H'' refers to methods that are fine-tuned on the Sintel, KITTI, and HD1K datasets. The bold font stands for the best result and the underlined results ranks 2nd. $\dagger$ corresponds to the results using a 3-step phantom gradient~\cite{PhantomGrad}. DEQ flow achieves SOTA zero-shot generalization results even w/o attention.
}
\vspace{-0.2cm}
\label{Tab:main_results}
\end{table*}

\begin{figure*}[!t]
    \centering
    \begin{overpic}[width=\linewidth]{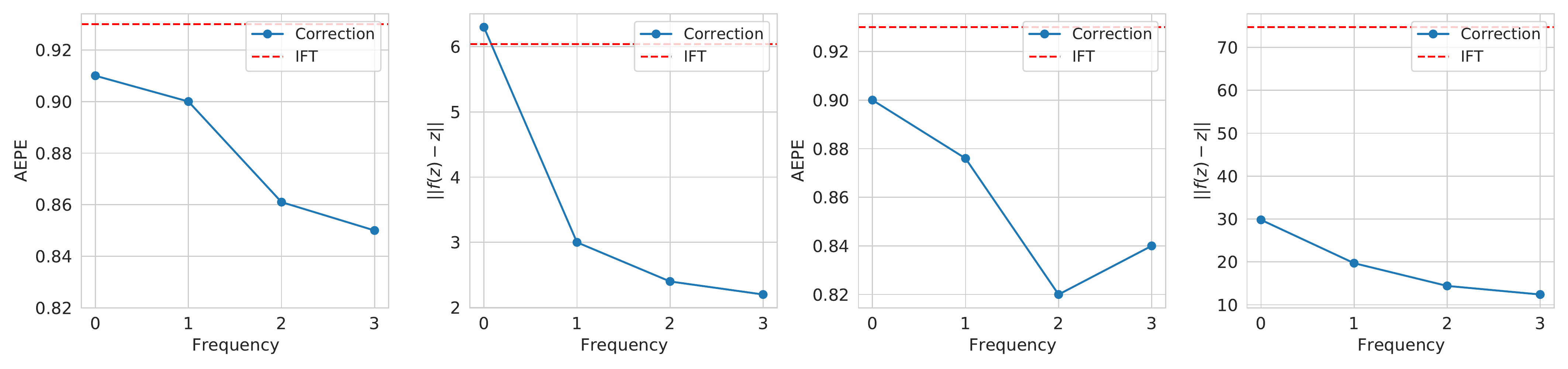}
        \put(14,-1.2){\scriptsize{(a) Training DEQ with Anderson forward solver}}
        \put(63.0,-1.2){\scriptsize{(b) Training DEQ with Broyden forward solver}}
        \put(11.4,23.3){\scriptsize{Performance}}
        \put(37.2,23.3){\scriptsize{Stability}}
        \put(61.15,23.3){\scriptsize{Performance}}
        \put(86.8,23.3){\scriptsize{Stability}}
    \end{overpic}
    \vspace{-3mm}    \caption{Performance and convergence stability (measured by absolute residual error) of the DEQ flow. Frequency indicates how many correction terms we pick, with $0$ meaning no correction. See the comparison with Jacobian Regularization~\cite{DEQ_JR} in the Appendix. DEQ flows trained with our proposed correction enjoy superior performance and stability.}
    \label{fig:correct}
    \vspace{-6mm}
\end{figure*}

We present the results of our experiments in this section. Specifically, we highlight the computational and memory efficiency of DEQ flow estimators and analyze how the fixed-point correction improves the DEQ flow.
Our method achieves \sota zero-shot performance on both the MPI Sintel~\cite{sintel} dataset and the KITTI 2015~\cite{kitti} dataset, with an astonishing $21.0\%$ error reduction in the F1-all measure and $14.9\%$ improvement in EPE for KITTI-15 (while still using a similar training budget to RAFT~\cite{RAFT}).

\subsection{Results}
\label{sec:results}

Our quantitative evaluation is presented in \cref{Tab:main_results}.
Following previous work~\cite{RAFT,GMA}, we first pretrain the DEQ flow model on the FlyingChairs~\cite{flownet} and FlyingThings3D~\cite{mayer2016large} datasets. We then test the model on the training set of MPI Sintel~\cite{sintel} and KITTI 2015~\cite{kitti} datasets. This model is denoted ``C + T''; it evaluates the \textit{zero-shot generalization} of the DEQ flow model. 
Then, we fine-tune the DEQ flow estimator on FlyingThings3D~\cite{mayer2016large}, MPI-Sintel~\cite{sintel}, KITTI 2015~\cite{kitti}, and HD1K~\cite{hd1k} for the test submission.

\begin{figure}[t]
\centering
\includegraphics[width=.43\textwidth]{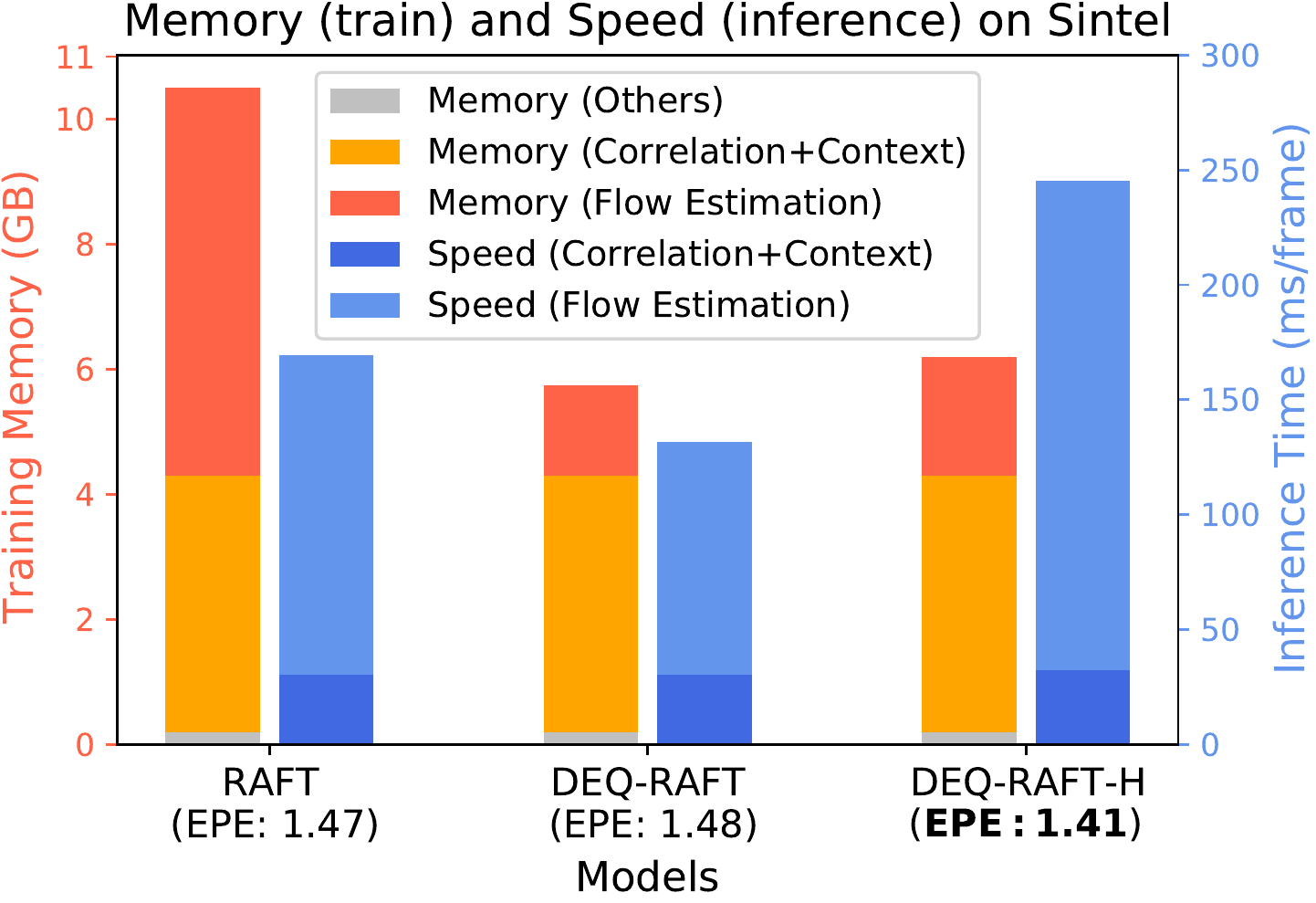}
\vspace{-.2cm}
\caption{Comparing the training memory, inference speed and performance on Sintel (clean) with image size $436 \times 1024$. The same model design (based on RAFT) consumes much less memory and computes much quicker than the recurrent counterpart. All results are benchmarked on a single Quadro RTX 8000 GPU.}
\label{fig:cost-comparison}
\vspace{-.4cm}
\end{figure}

The models are of exactly the same size as RAFT~(5.3M)~\citep{RAFT} and GMA~(5.9M)~\citep{GMA} except they use DEQ flow formulation instead of recurrent updates.
They are denoted as DEQ-RAFT-B and DEQ-GMA-B, respectively.
Exploiting the memory efficiency of the DEQ flow model (see \cref{subsec:performance-compute-tradeoff}), we can fit much larger models into the same compute budget of two 11~GB 2080Ti GPUs.
To this end, we also trained DEQ-RAFT-L~(8.4M) and DEQ-RAFT-H~(12.8M) by increasing the the width of hidden layers inside the equilibrium module $\f$.
As shown in \cref{fig:cost-comparison}, even the largest DEQ-RAFT-H model only consumes less than half of the flow estimation memory used by a standard-sized RAFT model, while achieving significantly better accuracy (4.38 AEPE and 14.9 F1-all score on KITTI-15, see \cref{Tab:main_results}).

\subsection{Performance-Compute Tradeoff}
\label{subsec:performance-compute-tradeoff}

We further verify the aforementioned computational and memory benefits of the DEQ flow model on the Sintel (clean)~\citep{sintel} dataset with a RAFT-based update operator (see Eq.~(\ref{eq:deq-flow-raft})) trained on FlyingChairs~\citep{flownet} and FlyingThings3D~\citep{mayer2016large}. The results are shown in Fig.~\ref{fig:cost-comparison}. Specifically, when training the DEQ flow estimator on Sintel with a batch size of 3 per GPU (the maximum that RAFT can fit with a 11~GB GPU), we observe that the memory cost of the flow estimation process reduces by a factor of over $4 \times$ (\textcolor{red}{red} bars). Note that since we keep the rest of the model intact (e.g., correlation pyramid and context extraction; see Sec.~\ref{subsec:preliminaries}), the DEQ flow estimator does not improve those parts of the memory burden, which now becomes the new dominant source of memory overhead. In addition, when we use the model for inference, we follow~\citet{RAFT} using 32 recurrent steps for RAFT (with warm-start), and the Anderson solver for DEQ-RAFT (with reuse), which stops if relative residual falls below $\varepsilon=10^{-3}$. Our results suggest that the DEQ flow converges to an accurate solution, and it is in practice about 20\% faster than the RAFT models with the same structure and size (\textcolor{blue}{blue} bars). Finally, we show that we can exploit such memory savings to build even larger and more accurate flow estimators (DEQ-RAFT-H), while still staying well within the compute and memory budget.

\vspace{-0.1cm}
\subsection{Ablation Study}

In this subsection, we aim to answer the following questions: 1) How useful is the fixed-point correction compared with canonical IFT in performance, stability, and speed? 2) How does the convergence of a DEQ flow correlate with the quality of the flow estimation? As in \cref{subsec:performance-compute-tradeoff}, we use the model design from RAFT~\cite{RAFT} to instantiate our DEQ flow. 
By default, we conduct the ablation experiments on the FlyingChairs~\cite{flownet} dataset using the default training hyperparameters of RAFT and report the Average End Point Error~(AEPE) on its validation split. 

\vspace{-0.3cm}
\paragraph{Stabilizing DEQ by Fixed-Point Correction.} As mentioned in Sec.~\ref{subsec:accelerate-deq-flow}, unregularized canonical DEQ models (as well as other implicit networks like Neural ODEs~\citep{chen2018neural}) typically suffer from a growing instability issue typically symptomized by an increasingly costly forward fixed-point solving process. We perform an ablation experiment to study how our proposed sparse fixed-point correction scheme could help alleviate this issue.
To understand the scheme's effect, we train a DEQ flow model using both an Anderson~\cite{anderson1965} and a Broyden~\cite{broyden1965class} solver with 36 and 24 forward iterations, respectively. For simplicity, we equally divide the solver convergence trajectory into $r+1$ segments (where $r$ is the frequency in \cref{fig:correct}) and impose a correction loss after each trajectory clip. As mentioned in Sec.~\ref{subsec:accelerate-deq-flow}, we apply the 1-step gradient~\citep{Ham,SamyFPN,PhantomGrad} to the correction loss. 

We visualize results of DEQ flow models trained with 3 different settings: 1) a DEQ flow trained by IFT directly without an auxiliary correction loss; 2) a DEQ flow trained by 1-step gradient without an auxiliary correction loss; and 3) DEQ flows trained by 1-step gradient \emph{as well as} 1-3 fixed-point correction terms. Our results are reported in terms of AEPE (which measures performance) and absolute fixed-point residual error $\|\f(\zstar; \x) - \zstar\|_2$ (which measures stability).
As shown in Fig.~\ref{fig:correct}, our proposed fixed-point correction significantly outperforms the standard IFT training protocol by about $9\%$, and reduces the fixed-point error by a conspicuous margin, \eg over $60\%$. Moreover, we find that the significant improvement in stability quickly diminishes as we apply more corrections, which suggests a sparse correction scheme.
Together with the inexact 1-step gradient, the total training time can be streamlined over $45\%$ compared with the IFT training schedule, while the backward pass of a DEQ flow is still almost \textit{free}.

\begin{figure}
\label{fig:corr}
\centering
\includegraphics[width=.45\textwidth]{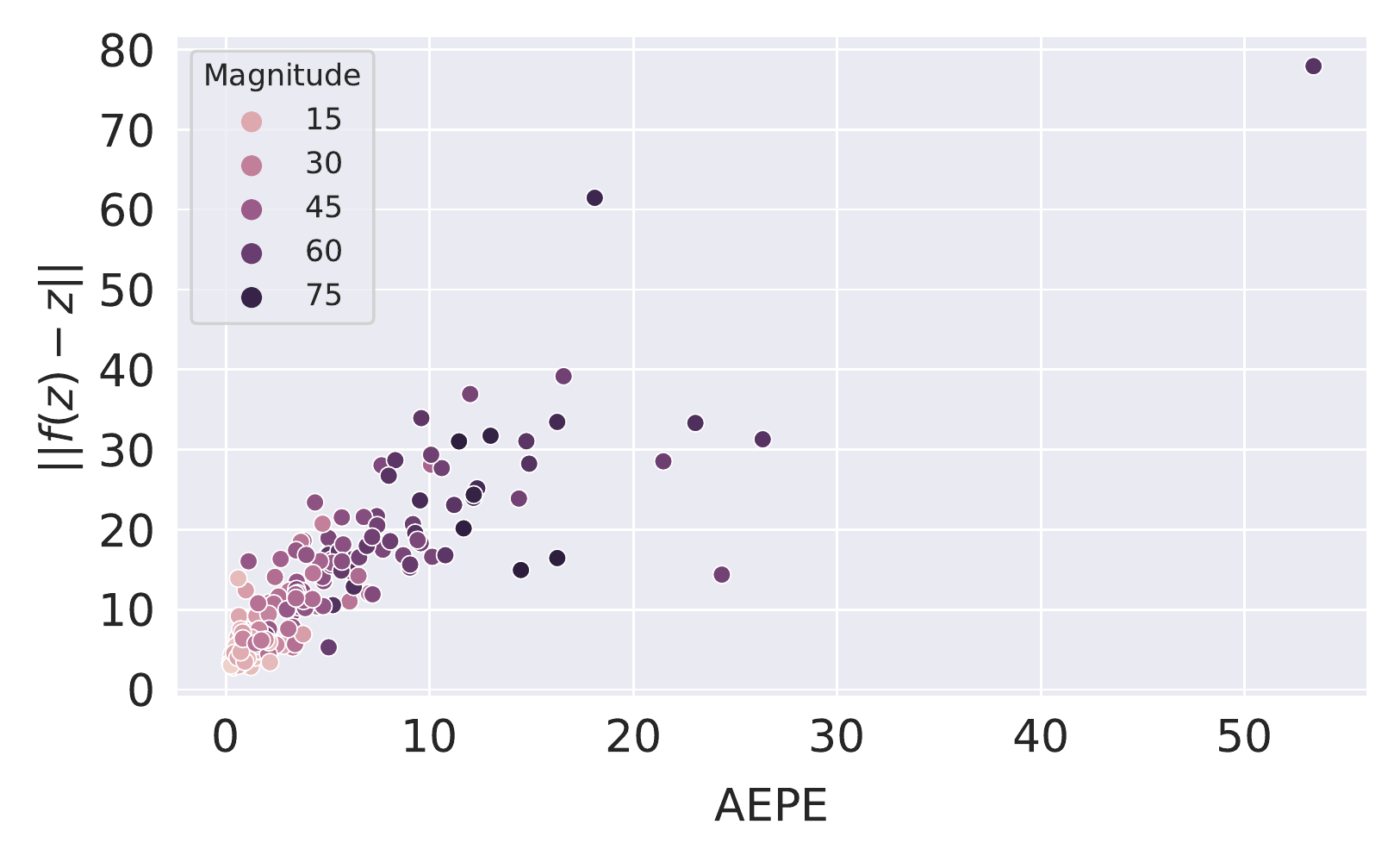}
\vspace{-.4cm}
\caption{Correlation between convergence and performance. We also observe that harder examples (e.g., those with large motion) typically lead to more challenging fixed-point convergence.}
\vspace{-.4cm}
\end{figure}

\vspace{-.3cm}
\paragraph{Correlation between Performance and Convergence.} A potential question is whether better fixed-point convergence can lead to better performance. To tackle this, we evaluate the DEQ flow model trained using the standard ``C+T'' training protocol (see \cref{sec:results}) on the KITTI-15~\cite{kitti} training set. We visualize the per-frame EPE and the convergence (measured by the absolute fixed point error) in \cref{fig:corr} and dye the scatter plot with the average norm of per-pixel flow across the frame, which can be understood as an indicator of hardness due to the large displacements. 
\begin{figure}
    \centering
    \includegraphics[width=7cm]{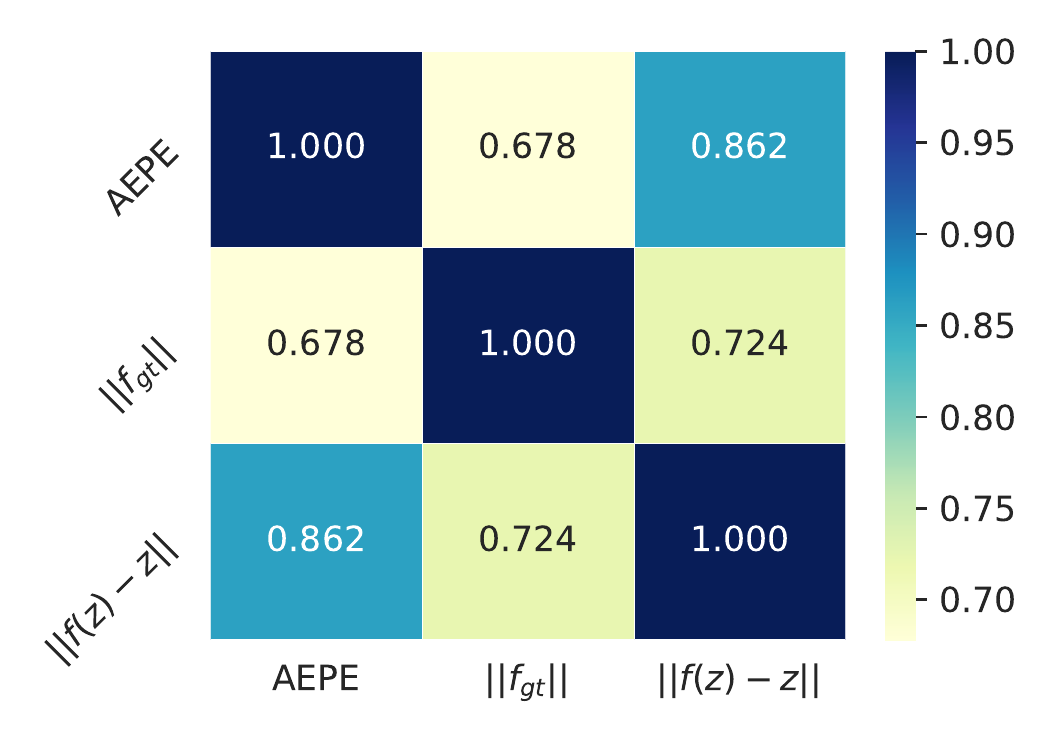}
    \vspace{-0.3cm}
    \caption{Pearson correlation coefficient across per-frame EPE, fixed-point error, and the magnitude of flow.}
    \label{fig:pearson}
    \vspace{-0.4cm}
\end{figure}
The Pearson correlation coefficient between the fixed-point error and EPE is over \textcolor{SeaGreen}{\textbf{0.86}} (see \cref{fig:pearson}) supporting the claim that convergence is strongly correlated with the flow performance. From \cref{fig:pearson}, we see that hard flows with large motions are also challenging for a naive solver. This demonstrates the necessity of advanced solvers in DEQ flow estimation.

\vspace{-0.2cm}
\section{Limitations}
The improved performance and efficiency of our approach comes at the cost of a slightly more complex training pipeline. 
Implementing the na\"ive unrolled flow estimation as presented in \citet{RAFT} and \citet{GMA} is simple using most libraries equipped with automatic differentiation that directly handle BPTT.
On the other hand, our approach involves some finagling of the training protocol (\eg fixed-point solvers, IFT, inexact gradients, \etc.). To help alleviate this complexity and promote the use of DEQ flows, we release our code at \url{https://github.com/locuslab/deq-flow}.

In addition, while DEQ flows provide a novel and more efficient framework to train and use these flow estimators, we still occasionally need to be careful about the stability of this approach. For example, what would happen if the solver \emph{converges poorly} (or even diverges) on a dynamical system? In such case, the behavior of the DEQ flow estimation would not be well-defined. In practice, we rarely observe such instability (as long as we spend enough solver steps); but as we analyzed in Sec. 4, harder examples also typically lead to more lengthy convergence path. We leave a more thorough study of estimation stability to future work.

\vspace{-0.2cm}
\section{Conclusion}
\label{sec:conclusion}

In this work, we introduce a new framework for modeling optical flow estimation. A deep equilibrium (DEQ) flow directly models and solves a fixed-point \emph{stable} flow estimate, and offers a set of tools that make these flow models' training and inference process highly efficient (e.g., they enjoy an almost-free backward pass). Moreover, the use of such equilibrium formulation is largely orthogonal to, and thus complements, the prior modeling and data efforts. We empirically show that it is possible to integrate the DEQ flow estimator with these model designs and achieve better performance on realistic optical flow datasets.
This implicit framework provides a strong (drop-in) replacement for existing recurrent update operators used by most cutting-edge flow estimators. The DEQ flows are both more powerful and lightweight --- both computationally and memory-wise. We believe this suggests an exciting direction for building more efficient, large-scale and accurate flow models in the future.

{\small
\bibliographystyle{unsrtnat}
\bibliography{deq}

\begin{thebibliography}{60}
\providecommand{\natexlab}[1]{#1}
\providecommand{\url}[1]{\texttt{#1}}
\expandafter\ifx\csname urlstyle\endcsname\relax
  \providecommand{\doi}[1]{doi: #1}\else
  \providecommand{\doi}{doi: \begingroup \urlstyle{rm}\Url}\fi

\bibitem[Teed and Deng(2020)]{RAFT}
Zachary Teed and Jia Deng.
\newblock Raft: Recurrent all-pairs field transforms for optical flow.
\newblock In \emph{European conference on computer vision}, pages 402--419.
  Springer, 2020.

\bibitem[Bai et~al.(2019)Bai, Kolter, and Koltun]{DEQ}
Shaojie Bai, J.~Zico Kolter, and Vladlen Koltun.
\newblock Deep equilibrium models.
\newblock In \emph{{Neural Information Processing Systems (NeurIPS)}}, 2019.

\bibitem[Jiang et~al.(2021{\natexlab{a}})Jiang, Campbell, Lu, Li, and
  Hartley]{GMA}
Shihao Jiang, Dylan Campbell, Yao Lu, Hongdong Li, and Richard Hartley.
\newblock Learning to estimate hidden motions with global motion aggregation.
\newblock In \emph{Proceedings of the IEEE/CVF International Conference on
  Computer Vision (ICCV)}, pages 9772--9781, October 2021{\natexlab{a}}.

\bibitem[Lucas et~al.(1981)Lucas, Kanade, et~al.]{lucas1981iterative}
Bruce~D Lucas, Takeo Kanade, et~al.
\newblock An iterative image registration technique with an application to
  stereo vision.
\newblock 1981.

\bibitem[Horn and Schunck(1981)]{horn1981determining}
Berthold~KP Horn and Brian~G Schunck.
\newblock Determining optical flow.
\newblock \emph{Artificial intelligence}, 17\penalty0 (1-3):\penalty0 185--203,
  1981.

\bibitem[Dosovitskiy et~al.(2015)Dosovitskiy, Fischer, Ilg, Hausser, Hazirbas,
  Golkov, Van Der~Smagt, Cremers, and Brox]{flownet}
Alexey Dosovitskiy, Philipp Fischer, Eddy Ilg, Philip Hausser, Caner Hazirbas,
  Vladimir Golkov, Patrick Van Der~Smagt, Daniel Cremers, and Thomas Brox.
\newblock Flownet: Learning optical flow with convolutional networks.
\newblock In \emph{Proceedings of the IEEE international conference on computer
  vision}, pages 2758--2766, 2015.

\bibitem[Zhao et~al.(2020)Zhao, Sheng, Dong, Chang, Xu, et~al.]{maskflownet}
Shengyu Zhao, Yilun Sheng, Yue Dong, Eric~I Chang, Yan Xu, et~al.
\newblock Maskflownet: Asymmetric feature matching with learnable occlusion
  mask.
\newblock In \emph{Proceedings of the IEEE/CVF Conference on Computer Vision
  and Pattern Recognition}, pages 6278--6287, 2020.

\bibitem[Hui et~al.(2019)Hui, Tang, and Loy]{liteflownet2}
Tak-Wai Hui, Xiaoou Tang, and Chen~Change Loy.
\newblock A lightweight optical flow cnn--revisiting data fidelity and
  regularization.
\newblock \emph{arXiv preprint arXiv:1903.07414}, 2019.

\bibitem[Cho et~al.(2014)Cho, {van Merrienboer}, Gulcehre, Bougares, Schwenk,
  and Bengio]{GRU}
Kyunghyun Cho, B~{van Merrienboer}, Caglar Gulcehre, F~Bougares, H~Schwenk, and
  Yoshua Bengio.
\newblock Learning phrase representations using rnn encoder-decoder for
  statistical machine translation.
\newblock In \emph{Conference on Empirical Methods in Natural Language
  Processing (EMNLP 2014)}, 2014.

\bibitem[Sun et~al.(2021)Sun, Vlasic, Herrmann, Jampani, Krainin, Chang, Zabih,
  Freeman, and Liu]{autoflow}
Deqing Sun, Daniel Vlasic, Charles Herrmann, V.~Jampani, Michael Krainin,
  Huiwen Chang, Ramin Zabih, William~T. Freeman, and Ce~Liu.
\newblock Autoflow: Learning a better training set for optical flow.
\newblock \emph{2021 IEEE/CVF Conference on Computer Vision and Pattern
  Recognition (CVPR)}, pages 10088--10097, 2021.

\bibitem[Eldesokey and Felsberg(2021)]{eldesokey2021normalized}
Abdelrahman Eldesokey and Michael Felsberg.
\newblock Normalized convolution upsampling for refined optical flow
  estimation.
\newblock \emph{arXiv preprint arXiv:2102.06979}, 2021.

\bibitem[Jiang et~al.(2021{\natexlab{b}})Jiang, Lu, Li, and
  Hartley]{Jiang2021LearningOF}
Shihao Jiang, Yao Lu, Hongdong Li, and Richard~I. Hartley.
\newblock Learning optical flow from a few matches.
\newblock \emph{2021 IEEE/CVF Conference on Computer Vision and Pattern
  Recognition (CVPR)}, pages 16587--16595, 2021{\natexlab{b}}.

\bibitem[Werbos(1990)]{werbos1990backpropagation}
Paul~J Werbos.
\newblock Backpropagation through time: what it does and how to do it.
\newblock \emph{Proceedings of the IEEE}, 78\penalty0 (10):\penalty0
  1550--1560, 1990.

\bibitem[Bai et~al.(2020)Bai, Koltun, and Kolter]{MDEQ}
Shaojie Bai, Vladlen Koltun, and J.~Zico Kolter.
\newblock {Multiscale Deep Equilibrium Models}.
\newblock In \emph{{Neural Information Processing Systems (NeurIPS)}}, pages
  5238--5250, 2020.

\bibitem[Winston and Kolter(2020)]{MON}
Ezra Winston and J.~Zico Kolter.
\newblock Monotone operator equilibrium networks.
\newblock In \emph{{Neural Information Processing Systems (NeurIPS)}}, pages
  10718--10728, 2020.

\bibitem[Kawaguchi(2020)]{kawaguchi2020theory}
Kenji Kawaguchi.
\newblock {On the Theory of Implicit Deep Learning: Global Convergence with
  Implicit Layers}.
\newblock In \emph{{International Conference on Learning Representations
  (ICLR)}}, 2020.

\bibitem[Kolter et~al.(2020)Kolter, Duvenaud, and Johnson]{Kolter2020}
J.~Zico Kolter, David Duvenaud, and Matthew Johnson.
\newblock Deep implicit layers tutorial - neural {ODEs}, deep equilibirum
  models, and beyond.
\newblock \emph{Neural Information Processing Systems Tutorial}, 2020.

\bibitem[Broyden(1965)]{broyden1965class}
Charles~G Broyden.
\newblock {A Class of Methods for Solving Nonlinear Simultaneous Equations}.
\newblock \emph{Mathematics of computation}, 19\penalty0 (92):\penalty0
  577--593, 1965.

\bibitem[Anderson(1965)]{anderson1965}
Donald~G. Anderson.
\newblock Iterative procedures for nonlinear integral equations.
\newblock \emph{Journal of the ACM (JACM)}, 12\penalty0 (4):\penalty0
  547–560, October 1965.

\bibitem[Fleet and Weiss(2006)]{fleet2006optical}
David Fleet and Yair Weiss.
\newblock Optical flow estimation.
\newblock In \emph{Handbook of mathematical models in computer vision}, pages
  237--257. Springer, 2006.

\bibitem[Geng et~al.(2021{\natexlab{a}})Geng, Guo, Chen, Li, Wei, and Lin]{Ham}
Zhengyang Geng, Meng-Hao Guo, Hongxu Chen, Xia Li, Ke~Wei, and Zhouchen Lin.
\newblock {Is Attention Better Than Matrix Decomposition?}
\newblock In \emph{{International Conference on Learning Representations
  (ICLR)}}, 2021{\natexlab{a}}.

\bibitem[Fung et~al.(2021)Fung, Heaton, Li, McKenzie, Osher, and Yin]{SamyFPN}
Samy~Wu Fung, Howard Heaton, Qiuwei Li, Daniel McKenzie, Stanley~J. Osher, and
  Wotao Yin.
\newblock {Fixed Point Networks: Implicit Depth Models with Jacobian-Free
  Backprop}.
\newblock \emph{arXiv preprint arXiv:2103.12803}, 2021.

\bibitem[Geng et~al.(2021{\natexlab{b}})Geng, Zhang, Bai, Wang, and
  Lin]{PhantomGrad}
Zhengyang Geng, Xin-Yu Zhang, Shaojie Bai, Yisen Wang, and Zhouchen Lin.
\newblock On training implicit models.
\newblock \emph{ArXiv}, abs/2111.05177, 2021{\natexlab{b}}.

\bibitem[Aleotti et~al.(2021)Aleotti, Poggi, and
  Mattoccia]{aleotti2021learning}
Filippo Aleotti, Matteo Poggi, and Stefano Mattoccia.
\newblock Learning optical flow from still images.
\newblock In \emph{Proceedings of the IEEE/CVF Conference on Computer Vision
  and Pattern Recognition}, pages 15201--15211, 2021.

\bibitem[Chen et~al.(2018)Chen, Rubanova, Bettencourt, and
  Duvenaud]{chen2018neural}
Tian~Qi Chen, Yulia Rubanova, Jesse Bettencourt, and David~K Duvenaud.
\newblock Neural ordinary differential equations.
\newblock In \emph{{Neural Information Processing Systems (NeurIPS)}}, 2018.

\bibitem[Bai et~al.(2021)Bai, Koltun, and Kolter]{DEQ_JR}
Shaojie Bai, Vladlen Koltun, and J.~Zico Kolter.
\newblock {Stabilizing Equilibrium Models by Jacobian Regularization}.
\newblock In \emph{{International Conference on Machine Learning (ICML)}},
  2021.

\bibitem[Geiger et~al.(2013)Geiger, Lenz, Stiller, and Urtasun]{kitti}
Andreas Geiger, Philip Lenz, Christoph Stiller, and Raquel Urtasun.
\newblock Vision meets robotics: The kitti dataset.
\newblock \emph{The International Journal of Robotics Research}, 32\penalty0
  (11):\penalty0 1231--1237, 2013.

\bibitem[Black and Anandan(1993)]{black1993framework}
Michael~J Black and Padmanabhan Anandan.
\newblock A framework for the robust estimation of optical flow.
\newblock In \emph{1993 (4th) International Conference on Computer Vision},
  pages 231--236. IEEE, 1993.

\bibitem[Zach et~al.(2007)Zach, Pock, and Bischof]{Zach2007ADB}
Christopher Zach, Thomas Pock, and Horst Bischof.
\newblock A duality based approach for realtime tv-l1 optical flow.
\newblock In \emph{DAGM-Symposium}, 2007.

\bibitem[Wedel et~al.(2008)Wedel, Pock, Zach, Bischof, and
  Cremers]{Wedel2008AnIA}
Andreas Wedel, Thomas Pock, Christopher Zach, Horst Bischof, and Daniel
  Cremers.
\newblock An improved algorithm for tv-l 1 optical flow.
\newblock In \emph{Statistical and Geometrical Approaches to Visual Motion
  Analysis}, 2008.

\bibitem[Brox and Malik(2011)]{Brox2011LargeDO}
Thomas Brox and Jitendra Malik.
\newblock Large displacement optical flow: Descriptor matching in variational
  motion estimation.
\newblock \emph{IEEE Transactions on Pattern Analysis and Machine
  Intelligence}, 33:\penalty0 500--513, 2011.

\bibitem[Menze et~al.(2015)Menze, Heipke, and Geiger]{Menze2015DiscreteOF}
Moritz Menze, Christian Heipke, and Andreas Geiger.
\newblock Discrete optimization for optical flow.
\newblock In \emph{GCPR}, 2015.

\bibitem[Chen and Koltun(2016)]{Chen2016FullFO}
Qifeng Chen and Vladlen Koltun.
\newblock Full flow: Optical flow estimation by global optimization over
  regular grids.
\newblock \emph{2016 IEEE Conference on Computer Vision and Pattern Recognition
  (CVPR)}, pages 4706--4714, 2016.

\bibitem[Xu et~al.(2017)Xu, Ranftl, and Koltun]{Xu2017AccurateOF}
Jia Xu, Ren{\'e} Ranftl, and Vladlen Koltun.
\newblock Accurate optical flow via direct cost volume processing.
\newblock \emph{2017 IEEE Conference on Computer Vision and Pattern Recognition
  (CVPR)}, pages 5807--5815, 2017.

\bibitem[Sun et~al.(2018{\natexlab{a}})Sun, Yang, Liu, and Kautz]{pwcnet}
Deqing Sun, Xiaodong Yang, Ming-Yu Liu, and Jan Kautz.
\newblock Pwc-net: Cnns for optical flow using pyramid, warping, and cost
  volume.
\newblock In \emph{Proceedings of the IEEE Conference on Computer Vision and
  Pattern Recognition}, pages 8934--8943, 2018{\natexlab{a}}.

\bibitem[Sun et~al.(2018{\natexlab{b}})Sun, Yang, Liu, and Kautz]{pwcnet+}
Deqing Sun, Xiaodong Yang, Ming-Yu Liu, and Jan Kautz.
\newblock Models matter, so does training: An empirical study of cnns for
  optical flow estimation.
\newblock \emph{arXiv preprint arXiv:1809.05571}, 2018{\natexlab{b}}.

\bibitem[Hui et~al.(2018)Hui, Tang, and Change~Loy]{liteflownet}
Tak-Wai Hui, Xiaoou Tang, and Chen Change~Loy.
\newblock Liteflownet: A lightweight convolutional neural network for optical
  flow estimation.
\newblock In \emph{Proceedings of the IEEE conference on computer vision and
  pattern recognition}, pages 8981--8989, 2018.

\bibitem[Ilg et~al.(2017)Ilg, Mayer, Saikia, Keuper, Dosovitskiy, and
  Brox]{ilg2017flownet}
Eddy Ilg, Nikolaus Mayer, Tonmoy Saikia, Margret Keuper, Alexey Dosovitskiy,
  and Thomas Brox.
\newblock Flownet 2.0: Evolution of optical flow estimation with deep networks.
\newblock In \emph{Proceedings of the IEEE conference on computer vision and
  pattern recognition}, pages 2462--2470, 2017.

\bibitem[Yang and Ramanan(2019)]{vcn}
Gengshan Yang and Deva Ramanan.
\newblock Volumetric correspondence networks for optical flow.
\newblock In \emph{Advances in Neural Information Processing Systems}, pages
  793--803, 2019.

\bibitem[Ranjan and Black(2017)]{ranjan2017optical}
Anurag Ranjan and Michael~J Black.
\newblock Optical flow estimation using a spatial pyramid network.
\newblock In \emph{Proceedings of the IEEE conference on computer vision and
  pattern recognition}, pages 4161--4170, 2017.

\bibitem[Zhang et~al.(2021)Zhang, Woodford, Prisacariu, and
  Torr]{Zhang2021SepFlow}
Feihu Zhang, Oliver~J. Woodford, Victor~Adrian Prisacariu, and Philip~H.S.
  Torr.
\newblock Separable flow: Learning motion cost volumes for optical flow
  estimation.
\newblock In \emph{Proceedings of the IEEE/CVF International Conference on
  Computer Vision (ICCV)}, 2021.

\bibitem[Jiang et~al.(2021{\natexlab{c}})Jiang, Lu, Li, and
  Hartley]{jiang2021learning}
Shihao Jiang, Yao Lu, Hongdong Li, and Richard Hartley.
\newblock Learning optical flow from a few matches.
\newblock In \emph{{IEEE Conference on Computer Vision and Pattern Recognition
  (CVPR)}}, pages 16592--16600, 2021{\natexlab{c}}.

\bibitem[Amos and Kolter(2017)]{amos2017optnet}
Brandon Amos and J.~Zico Kolter.
\newblock {OptNet}: Differentiable optimization as a layer in neural networks.
\newblock In \emph{{International Conference on Machine Learning (ICML)}},
  2017.

\bibitem[{El Ghaoui} et~al.(2019){El Ghaoui}, {Gu}, {Travacca}, and
  {Askari}]{elghaoui2019implicit}
Laurent {El Ghaoui}, Fangda {Gu}, Bertrand {Travacca}, and Armin {Askari}.
\newblock Implicit deep learning.
\newblock \emph{arXiv:1908.06315}, 2019.

\bibitem[Dupont et~al.(2019)Dupont, Doucet, and Teh]{dupont2019augmented}
Emilien Dupont, Arnaud Doucet, and Yee~Whye Teh.
\newblock Augmented neural {ODEs}.
\newblock In \emph{{Neural Information Processing Systems (NeurIPS)}}, 2019.

\bibitem[Lu et~al.(2021)Lu, Chen, Li, Wang, and Zhu]{lu2021implicit}
Cheng Lu, Jianfei Chen, Chongxuan Li, Qiuhao Wang, and Jun Zhu.
\newblock Implicit normalizing flows.
\newblock In \emph{{International Conference on Learning Representations
  (ICLR)}}, 2021.

\bibitem[Vaswani et~al.(2017)Vaswani, Shazeer, Parmar, Uszkoreit, Jones, Gomez,
  Kaiser, and Polosukhin]{vaswani2017attention}
Ashish Vaswani, Noam Shazeer, Niki Parmar, Jakob Uszkoreit, Llion Jones,
  Aidan~N Gomez, {\L}ukasz Kaiser, and Illia Polosukhin.
\newblock Attention is all you need.
\newblock In \emph{{Neural Information Processing Systems (NeurIPS)}}, 2017.

\bibitem[He et~al.(2016)He, Zhang, Ren, and Sun]{he2016deep}
Kaiming He, Xiangyu Zhang, Shaoqing Ren, and Jian Sun.
\newblock Deep residual learning for image recognition.
\newblock In \emph{{IEEE Conference on Computer Vision and Pattern Recognition
  (CVPR)}}, 2016.

\bibitem[Gu et~al.(2020)Gu, Chang, Zhu, Sojoudi, and Ghaoui]{gu2020implicit}
Fangda Gu, Heng Chang, Wenwu Zhu, Somayeh Sojoudi, and Laurent~El Ghaoui.
\newblock Implicit graph neural networks.
\newblock \emph{arXiv preprint arXiv:2009.06211}, 2020.

\bibitem[Park et~al.(2021)Park, Choo, and Park]{park2021convergent}
Junyoung Park, Jinhyun Choo, and Jinkyoo Park.
\newblock Convergent graph solvers.
\newblock \emph{arXiv preprint arXiv:2106.01680}, 2021.

\bibitem[Liu et~al.(2021)Liu, Kawaguchi, Hooi, Wang, and Xiao]{liu2021eignn}
Juncheng Liu, Kenji Kawaguchi, Bryan Hooi, Yiwei Wang, and Xiaokui Xiao.
\newblock {EIGNN}: Efficient infinite-depth graph neural networks.
\newblock In A.~Beygelzimer, Y.~Dauphin, P.~Liang, and J.~Wortman Vaughan,
  editors, \emph{Advances in Neural Information Processing Systems}, 2021.

\bibitem[Krantz and Parks(2012)]{krantz2012implicit}
Steven~G Krantz and Harold~R Parks.
\newblock \emph{The implicit function theorem: History, theory, and
  applications}.
\newblock Springer, 2012.

\bibitem[Kelly et~al.(2020)Kelly, Bettencourt, Johnson, and
  Duvenaud]{kelly2020learning}
Jacob Kelly, Jesse Bettencourt, Matthew~James Johnson, and David Duvenaud.
\newblock Learning differential equations that are easy to solve.
\newblock In \emph{{Neural Information Processing Systems (NeurIPS)}}, 2020.

\bibitem[Hutchinson(1989{\natexlab{a}})]{hutchinson1989stochastic}
Michael~F Hutchinson.
\newblock A stochastic estimator of the trace of the influence matrix for
  laplacian smoothing splines.
\newblock \emph{Communications in Statistics-Simulation and Computation},
  18\penalty0 (3):\penalty0 1059--1076, 1989{\natexlab{a}}.

\bibitem[Butler et~al.(2012)Butler, Wulff, Stanley, and Black]{sintel}
Daniel~J Butler, Jonas Wulff, Garrett~B Stanley, and Michael~J Black.
\newblock A naturalistic open source movie for optical flow evaluation.
\newblock In \emph{European conference on computer vision}, pages 611--625.
  Springer, 2012.

\bibitem[Mayer et~al.(2016)Mayer, Ilg, Hausser, Fischer, Cremers, Dosovitskiy,
  and Brox]{mayer2016large}
Nikolaus Mayer, Eddy Ilg, Philip Hausser, Philipp Fischer, Daniel Cremers,
  Alexey Dosovitskiy, and Thomas Brox.
\newblock A large dataset to train convolutional networks for disparity,
  optical flow, and scene flow estimation.
\newblock In \emph{Proceedings of the IEEE conference on computer vision and
  pattern recognition}, pages 4040--4048, 2016.

\bibitem[Kondermann et~al.(2016)Kondermann, Nair, Honauer, Krispin, Andrulis,
  Brock, Gussefeld, Rahimimoghaddam, Hofmann, Brenner, et~al.]{hd1k}
Daniel Kondermann, Rahul Nair, Katrin Honauer, Karsten Krispin, Jonas Andrulis,
  Alexander Brock, Burkhard Gussefeld, Mohsen Rahimimoghaddam, Sabine Hofmann,
  Claus Brenner, et~al.
\newblock The hci benchmark suite: Stereo and flow ground truth with
  uncertainties for urban autonomous driving.
\newblock \emph{CVPR Workshop}, 2016.

\bibitem[Paszke et~al.(2019)Paszke, Gross, Massa, Lerer, Bradbury, Chanan,
  Killeen, Lin, Gimelshein, Antiga, Desmaison, Kopf, Yang, DeVito, Raison,
  Tejani, Chilamkurthy, Steiner, Fang, Bai, and Chintala]{pythorch}
Adam Paszke, Sam Gross, Francisco Massa, Adam Lerer, James Bradbury, Gregory
  Chanan, Trevor Killeen, Zeming Lin, Natalia Gimelshein, Luca Antiga, Alban
  Desmaison, Andreas Kopf, Edward Yang, Zachary DeVito, Martin Raison, Alykhan
  Tejani, Sasank Chilamkurthy, Benoit Steiner, Lu~Fang, Junjie Bai, and Soumith
  Chintala.
\newblock {PyTorch: An Imperative Style, High-performance Deep Learning
  Library}.
\newblock In \emph{{Neural Information Processing Systems (NeurIPS)}}, pages
  8026--8037, 2019.

\bibitem[Bai et~al.(2022)Bai, Koltun, and Kolter]{bai2022neural}
Shaojie Bai, Vladlen Koltun, and J~Zico Kolter.
\newblock Neural deep equilibrium solvers.
\newblock In \emph{International Conference on Learning Representations}, 2022.

\bibitem[Hutchinson(1989{\natexlab{b}})]{Hutchinson}
Michael~F Hutchinson.
\newblock A stochastic estimator of the trace of the influence matrix for
  laplacian smoothing splines.
\newblock \emph{Communications in Statistics-Simulation and Computation},
  18\penalty0 (3):\penalty0 1059--1076, 1989{\natexlab{b}}.

\end{thebibliography}
}

\clearpage
\appendix

\section{Pseudo Code}

We provide a PyTorch-style~\cite{pythorch} pseudo-code for the DEQ flow in \cref{alg:deq-torch-short}. Besides fixed-point reuse and an inexact (one-step) gradient as shown previously, we also include the fixed-point correction loss (applied with \texttt{freq}). In practice, we can set \texttt{freq}$=1$, and use either Broyden's method~\cite{broyden1965class} or Anderson acceleration~\cite{anderson1965} as \texttt{solver}. This sparse fixed-point correction scheme encourages stable training dynamics, which we analyze further in \cref{fig:jr-correction}.

\begin{algorithm}[t]
\caption{DEQ flow (PyTorch-style). Note that we reuse the fixed point and perform fixed-point correction.}
\label{alg:deq-torch-short}
\begin{lstlisting}[style=Pytorch,escapeinside={(@}{@)}]
# solver: fixed-point solver, e.g., Broyden(@~\cite{broyden1965class}@)
# func: layer (@$\f$@) that defines dynamic system
# dist: loss function for fixed point correction
# x: input information (@$\vx_t=(\mathbf{q}_t, \mathcal{C}_t)$@) of frame (@$t$@)
# z: fixed-point flow estimation (@$\vz_t^{*}$@)
# f: ground truth optical flow (@$\rvf$@)
# freq: frequency of correction
# gamma: coefficient of correction
# prev_z: (@$\vz_{t-1}^{*}$@) of the last frame (if exists)
# training: bool indicating training/inference

# Forward pass (w/ backward pass by autodiff)
def forward(x, f, gamma, freq=1, 
    training=True, prev_z=None):
    with torch.no_grad(): # Fixed-Point Reuse
        z, z_m = solver(func, x, freq, z0=prev_z) 
        
    if training:
        loss = dist(f, func(z, x))
        # Fixed Point Correction w/ 1-step gradient
        for i in range(freq):
            z_mi = func(z_m[i], x)
            loss = loss + gamma[i] * dist(f, z_mi)
        return loss
        
    return z
\end{lstlisting}
\vspace{-0.1cm}
\end{algorithm}

\section{Experiment Settings}
In this section, we present the detailed experiment settings for training and inference with the DEQ flow estimators. The code will be made publicly available upon acceptance.

\subsection{Model Design}

As mentioned previously, a deep equilibrium (DEQ) flow estimator subsumes a wide variety of model designs, and can be integrated with the latest, cutting-edge update operators. We show the integration of two of the most prominent designs that have achieved state-of-the-art optical flow results below, while noting in general that other alternatives are also possible.

\paragraph{DEQ flow by RAFT.}
Without any modification to the original design of RAFT~\cite{RAFT}, we can instantiate a DEQ-RAFT by defining the equilibrium system as follows, 
\begin{equation}
\begin{array}{llll}
& \x          & = & \text{Conv2d}\left([\q,\, \fstar, \gC(\fstar+\c^0)]\right) \\[0.3mm]
& \hstar      & = & \text{ConvGRU}\left(\hstar, [\x,\, \q] \right)   \\[0.15mm]
& \fstar      & = & \fstar + \text{Conv2d}\left(\hstar \right),        \\
\end{array} 
\end{equation}
where $\gC(\fstar+\c^0)$ stands for the correlation lookup as in RAFT~\cite{RAFT}, \text{Conv2d} stands for 2D convolutional layers with ReLU activations, and \text{ConvGRU} represents a GRU-style gated activation following convolutions, respectively. We refer the readers to~\citet{RAFT} and the code base\footnote{\href{https://github.com/princeton-vl/RAFT}{https://github.com/princeton-vl/RAFT}} for more details.

\paragraph{DEQ flow by GMA.}
More recently,~\citet{GMA} show that we can improve on the formulation of RAFT above by adding an attention module to better model the occlusion scenarios in video frames. Specifically, we also provide an instantiation of such Global Motion Aggregation~(GMA) update operator~\citep{GMA} in the context of DEQ flows, where we solve for the equilibrium $\zstar=(\hstar, \fstar)$ that satisfies
\begin{equation}
\label{eq:deq-flow-gma}
\begin{array}{llll}
& \x          & = & \text{Conv2d}\left([\q,\, \fstar, \gC(\fstar+\c^0)]\right)    \\[0.15mm]
& \hat{\x}    & = & \text{Attention}\left(\q,\, \q,\, \x\right)                   \\[0.15mm]
& \hstar      & = & \text{ConvGRU}\left(\hstar, [\hat{\x},\, \x,\, \q]\right)     \\[0.15mm]
& \fstar      & = & \fstar + \text{Conv2d}\left(\hstar \right)                    \\
\end{array} 
\end{equation}
where $\text{Attention}$ is the attention module, see~\cite{GMA,vaswani2017attention}.

\begin{figure*}[!t]
\centering
\vspace{0cm}
\begin{overpic}[width=\textwidth]{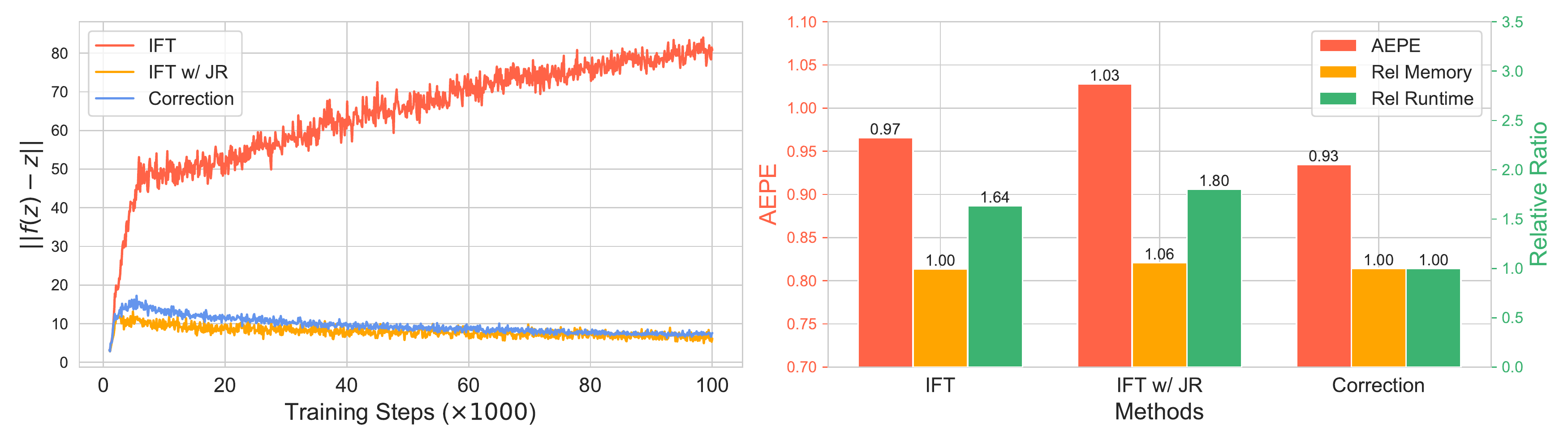}
        \put(17.5,-1.2){(a) Training stability}
        \put(54.0,-1.2){(b) Training cost and generalization performance}
    \end{overpic}
\vspace{-0.2cm}
\captionof{figure}{
\tbf{Comparison of IFT, Jacobian Regularization and Fixed-Point Correction.}
Given a limited forward solver budget, the fixed-point correction protocol successfully stabilizes training and shows accelerated fixed-point convergence with visible performance improvements over Jacobian Regularization~\cite{DEQ_JR} for DEQ. 
}
\vspace{0cm}
\label{fig:jr-correction}
\end{figure*}

\paragraph{Model Hyperparameters}
For the base models, DEQ-RAFT-B and DEQ-GMA-B, we employ the \emph{exact same architecture and hyperparameter choice} for the equilibrium module $\f$ as originally used by RAFT~\citep{RAFT} and GMA~\citep{GMA}. We merely replace the recurrent (and BPTT-based) formulation with a fixed-point system-based one (and the backward pass with IFT or inexact gradients). For the Large models and Huge models, \ie DEQ-RAFT-L, DEQ-GMA-L, and DEQ-RAFT-H, we use the same designs as the backbone models while increasing the number of hidden dimensions by a factor of $1.5\times$ and $2\times$, respectively.

\subsection{Training Details}

Following the settings of prior work~\cite{RAFT,GMA}, we apply a four-stage training protocol using the default hyperparameters unless specified otherwise. First, we train DEQ-RAFT-L, DEQ-RAFT-H, and DEQ-RAFT-H$^\dagger$ on FlyingChairs~\cite{flownet} for 120k iterations and then on FlyingThings~\cite{mayer2016large} for another 120k iterations. In addtion, we also train the DEQ-GMA instantiation on FlyingChairs~\cite{flownet} using a batch size of 10 and a learning rate of 4e-4, with the same setting as RAFT~\cite{RAFT}. We note that the original recurrent GMA model quickly exhausts the memory budget even with a much smaller batch size and mixed-precision training, which is not a concern for DEQ-GMA due to the implicit modeling framework.

We use an Anderson acceleration~\citep{anderson1965} solver with up to 40 forward steps for the base model (DEQ-RAFT-B).
For larger models (DEQ-RAFT-L, DEQ-RAFT-H, DEQ-GMA-B, and DEQ-GMA-L), we use a Broyden~\cite{broyden1965class} solver with up to 36 forward steps.
For the pretraining on Chairs, we introduce 1-3 correction terms accordingly.
We adopt the 1-step gradient~\cite{Ham,SamyFPN,PhantomGrad} for training, which suggests almost-free backward passes. 
All the above experiments can be conducted on two 11~GB GPUs (NVIDIA 2080Ti).

For DEQ-RAFT-H$^\dagger$, we employ a Broyden~\cite{broyden1965class} solver with 36 forward steps and 5 uniformly spaced correction terms. 
We apply phantom gradient~\cite{PhantomGrad} of 3 steps to DEQ-RAFT-H$^\dagger$ and average the best results of 3 runs under a 5k-step evaluation gap to report the ``C+T'' schedule performance.
The memory budget for this experiment can be two 16 GB GPUs or three 11 GB GPUs, while mixed-precision can further reduce the memory cost.

Note that we do not manually select the damping factor $\lambda$~\cite{PhantomGrad} but adopt the gating function from ConvGRU as the adaptive $\lambda$, which waives the hyperparameter tuning for the phantom gradient.
The fixed point solvers are based on that of~\citet{DEQ}. In all those cases, the fixed-point solvers stop either when the iterations reach the limit, or if the absolute residual error falls below $0.001$. 

Our results suggest that, with minimal tuning, these settings are sufficient to achieve \sota results (while imposing much lower compute and memory cost).
Empirically, we note that a DEQ flow model trained using a stronger solver (\eg the Broyden solver) and with more steps typically leads to slightly better performance. This implies the prospect of potentially further boosting the DEQ-flow performance by more precise fixed-point solving, while noting that recent progress on using neural solvers might suggest an interesting perspective~\citep{bai2022neural}.

\section{DEQ Flows with Fixed-point Correction}

The growing instability problem has been a longstanding challenge in training implicit neural networks like DEQs. One of the contributions of this paper is also the introduction of fixed-point correction term to stabilize the DEQ flow estimation convergence. Specifically, previous methods rely on more constrained regularization settings such as Jacobian-based losses~\citep{DEQ_JR}, \ie penalizing the upper bound of Jacobian spectral radius,
\begin{equation*}
    \rho(J_{\f}(\zstar)) \leq \| J_{\f}(\zstar) \|_F = \sqrt{\text{tr}(J_{\f}^\top J_{\f})},
\end{equation*}
where $J_{\f}(\zstar) \in \Rdim{d \times d}$ denotes the Jacobian of $\f$ at $\zstar$, $\rho$ corresponds to the spectral radius of a square matrix. By the stochastic Hutchinson trace estimator~\cite{Hutchinson}, we have 
\begin{equation*}
    \text{tr}(J_{\f}^\top J_{\f}) = \mathbb{E}_{\epsilon\sim p(\epsilon)} \left[\epsilon^\top J_{\f}^\top J_{\f} \epsilon \right] \approx \sum_{\epsilon\sim p(\epsilon)} \| J_{\f} \epsilon \|_2^2,
\end{equation*}
where $p(\epsilon)$ can be the Gaussian distribution $\mathcal{N}(0, I_d)$ or the Rademacher distribution.
Different from prior works, we advocate for exploiting the benefit of IFT and inexact gradient to \emph{sparsely} apply a fixed-point correction scheme to the convergence path. 

In this section, we present an ablation study on FlyingChairs~\cite{flownet} that compare the stability and generalization performance of DEQ flow models trained in three different settings: 1) standard implicit differentiatio (i.e., IFT); 2) standard IFT with Jacobian regularization~\citep{DEQ_JR}; and 3) our proposed fixed-point correction scheme with a single correction term.
As mentioned previously, we perform 1-step inexact gradient on the correction loss as well. For the purpose of this ablation, we run the forward fixed-point solver for a limited compute budget of 16 Anderson~\cite{anderson1965} steps in all three settings, and analyze their convergence behavior accordingly.

As shown in \cref{fig:jr-correction}, the model trained using the standard implicit function theorem (IFT) suffers from the ``growing instability'' issue (see \textcolor{tomato}{red} curve in \cref{fig:jr-correction} (a)), as described in prior works indeed~\citep{DEQ,DEQ_JR,chen2018neural,MON}. While strong enough Jacobian regularization can indeed stabilize the training process and lead to good overall convergence (see \textcolor{orange}{orange} curve in \cref{fig:jr-correction} (a)), we observe that it is usually at a heavy cost of optical flow estimation accuracy (see \cref{fig:jr-correction} (b)). This agrees with the conclusion of~\citet{DEQ_JR}. In contrast, we find it suffices to use a single fixed-point correction term in DEQ flow to achieve the same stabilizing effect (see \textcolor{cornflowerblue}{blue} curve in \cref{fig:jr-correction} (a)) \emph{at no extra cost} to the average EPE on the validation set. We hypothesize that such a fixed-point correction method may suggest an elegant and lightweight solution to the growing instability problem in the broader implicit deep learning community beyond the scope of optical flow estimation.

\section{Qualitative Results}

\begin{figure*}[!ht]
    \centering
    \begin{overpic}[width=0.24\linewidth]{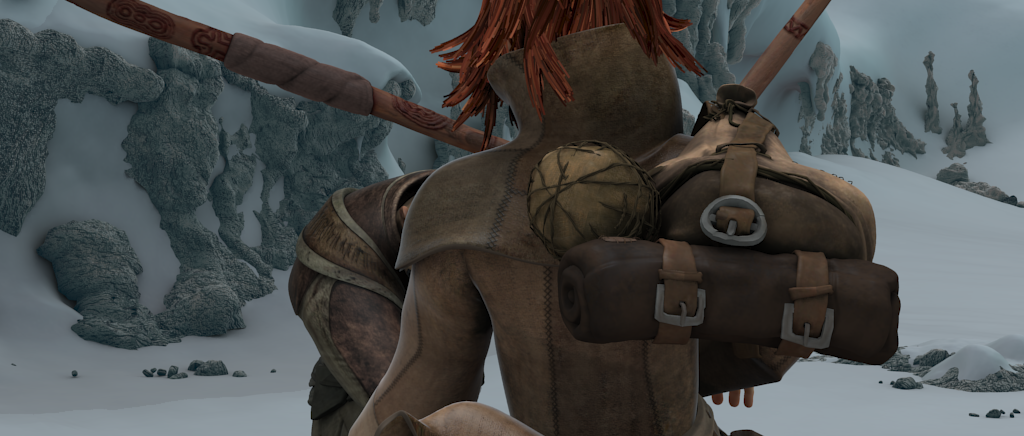}
    \end{overpic}%
    \begin{overpic}[width=0.24\linewidth]{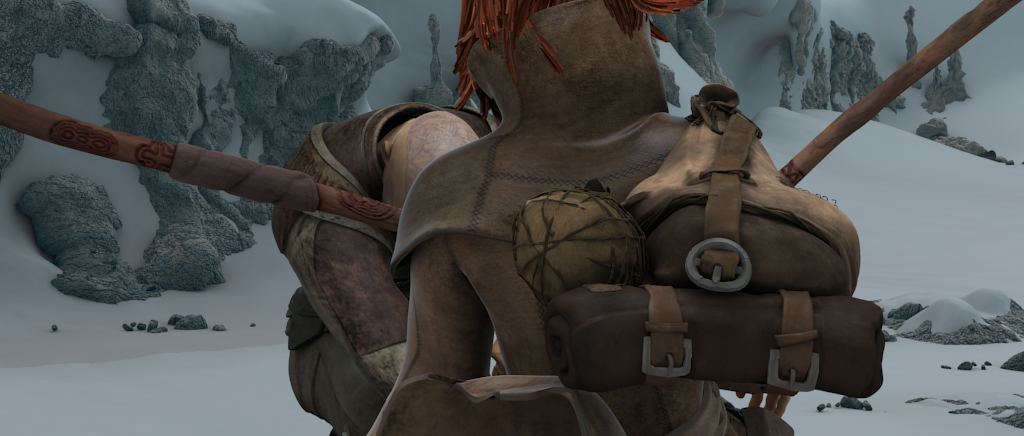}
    \end{overpic}%
    \begin{overpic}[width=0.24\linewidth]{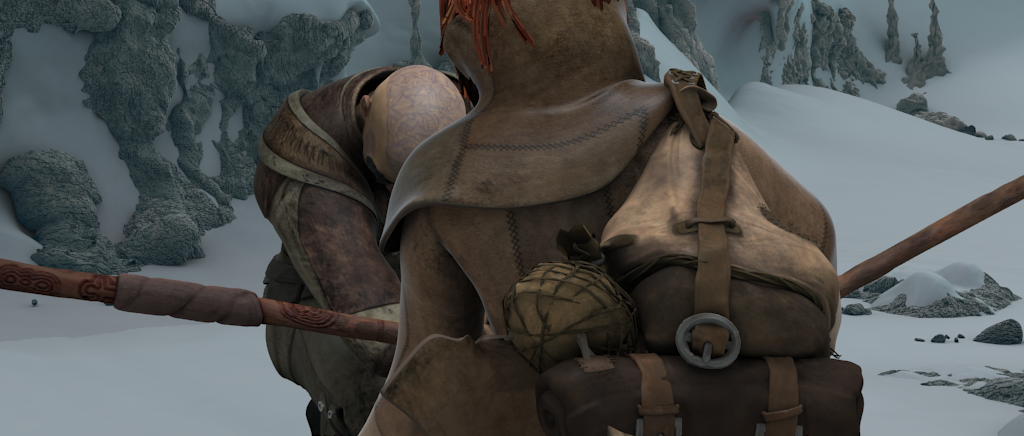}
    \end{overpic}%
    \begin{overpic}[width=0.24\linewidth]{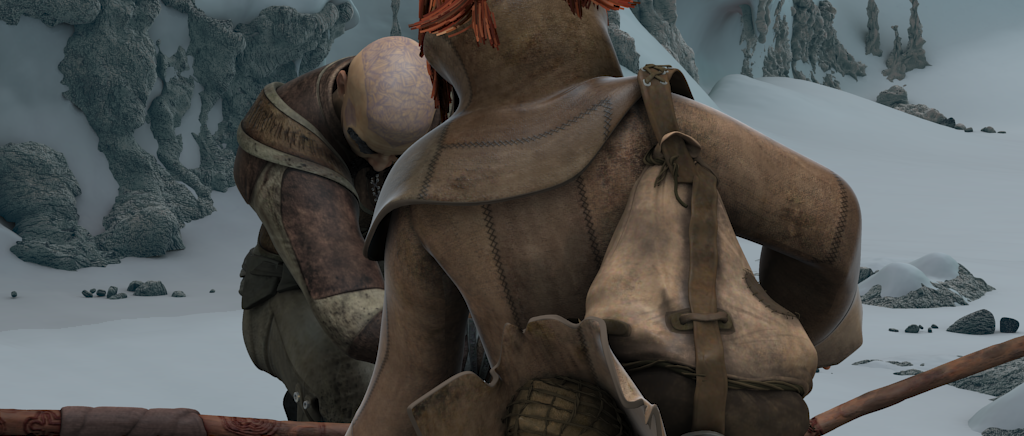}
    \end{overpic}%
    \\
    \begin{overpic}[width=0.24\linewidth]{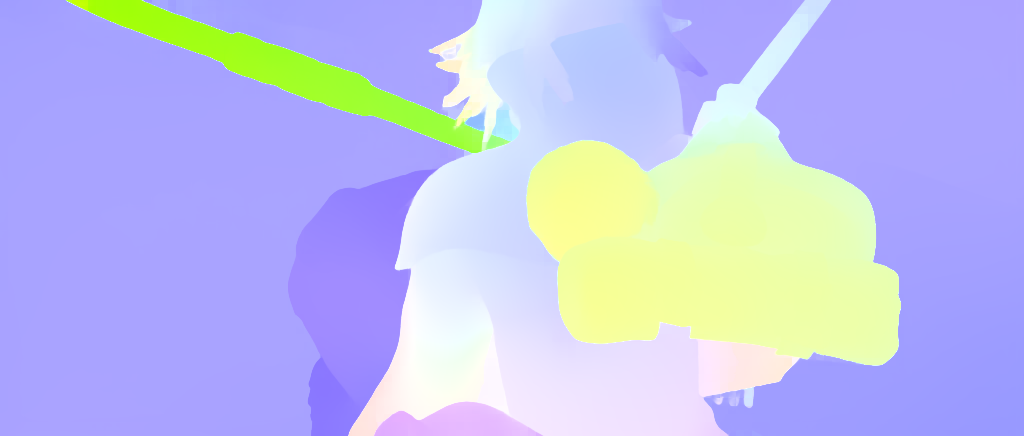}
        \put(34,-6.0){\scriptsize{(a) Frame 12}}
    \end{overpic}%
    \begin{overpic}[width=0.24\linewidth]{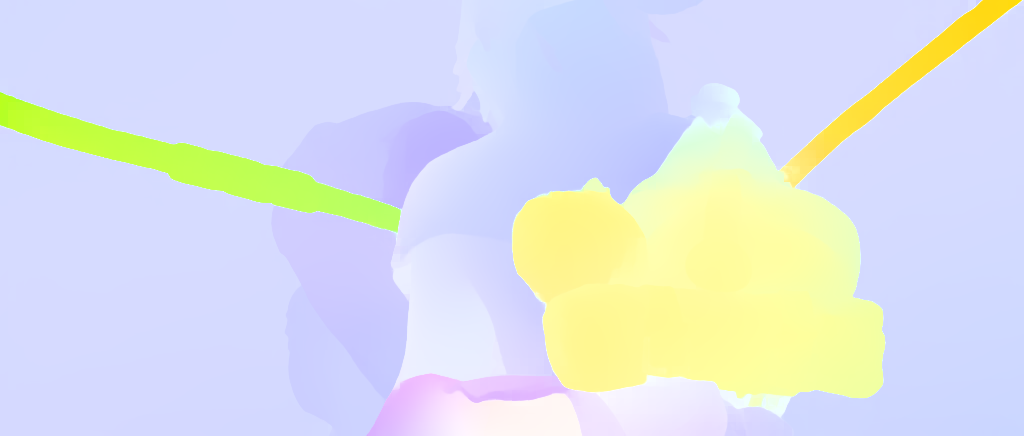}
        \put(34,-6.0){\scriptsize{(b) Frame 13}}
    \end{overpic}%
    \begin{overpic}[width=0.24\linewidth]{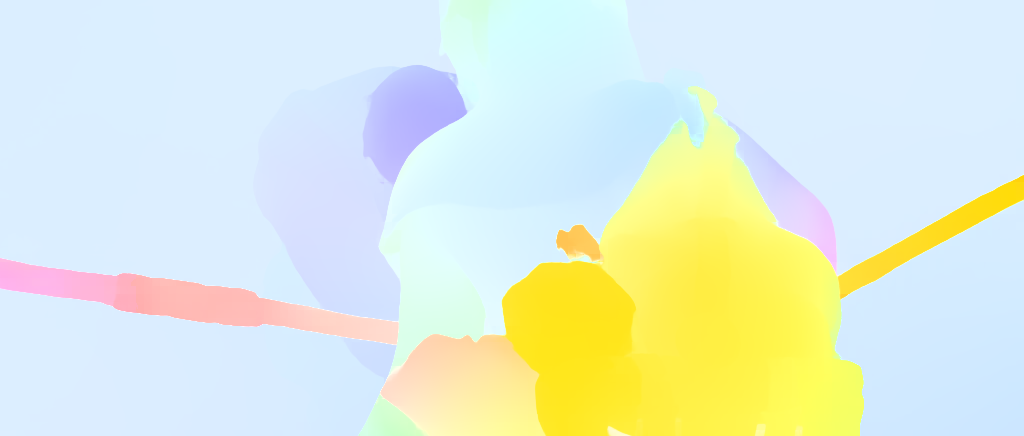}
        \put(34,-6.0){\scriptsize{(c) Frame 14}}
    \end{overpic}%
    \begin{overpic}[width=0.24\linewidth]{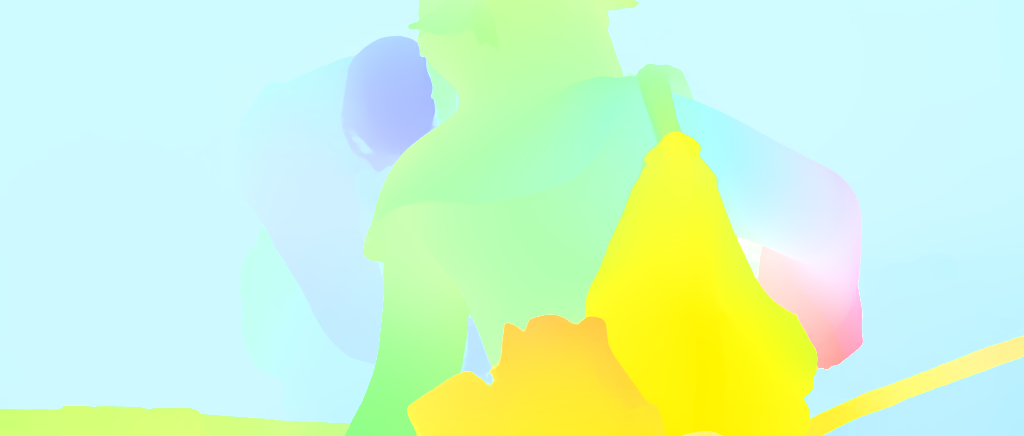}
        \put(34,-6.0){\scriptsize{(d) Frame 15}}
    \end{overpic}%
    \vspace{4.5pt}
    \caption{
      Visualization on the Sintel test set, \texttt{ambush\_1} sequence of the clean split.
    }
    \label{fig:demo-clean-ambush-1}
    \vspace{-6pt}
\end{figure*}

\begin{figure*}[!ht]
    \centering
    \begin{overpic}[width=0.24\linewidth]{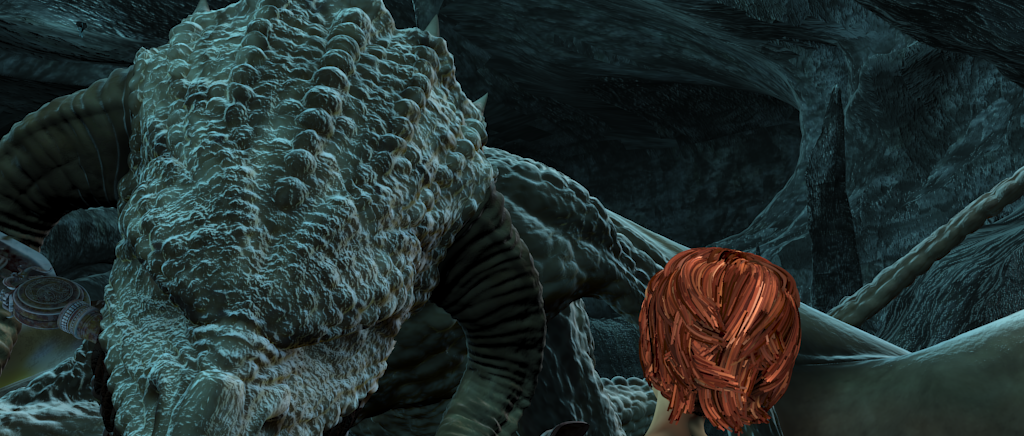}
    \end{overpic}%
    \begin{overpic}[width=0.24\linewidth]{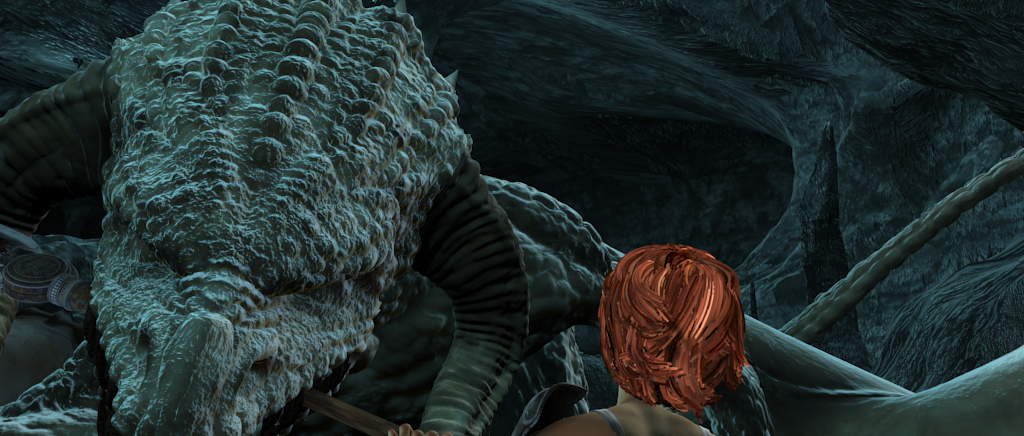}
    \end{overpic}%
    \begin{overpic}[width=0.24\linewidth]{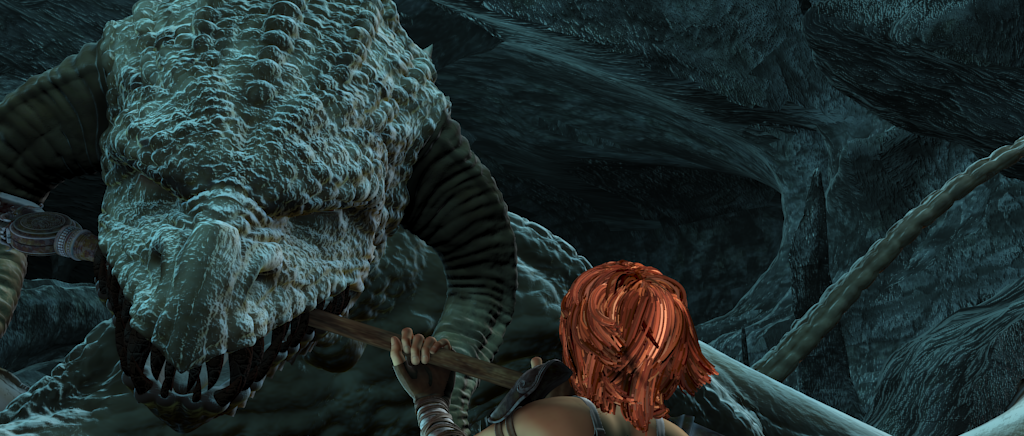}
    \end{overpic}%
    \begin{overpic}[width=0.24\linewidth]{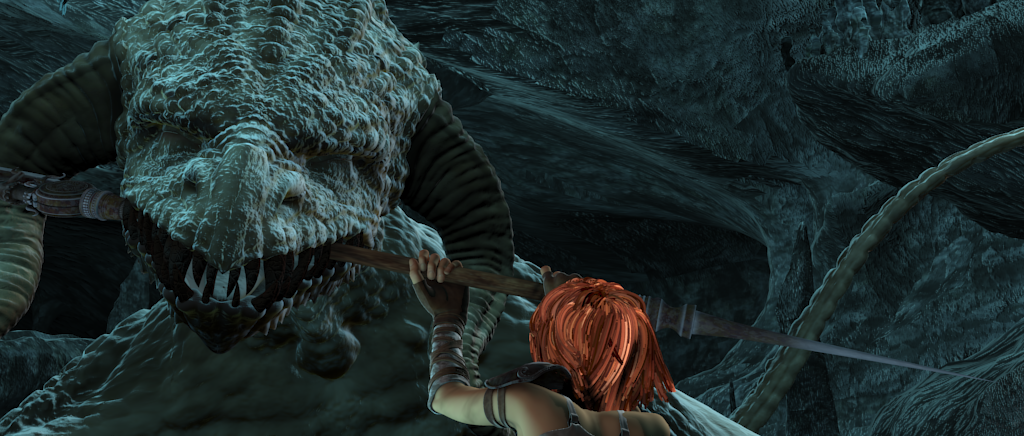}
    \end{overpic}%
    \\
    \begin{overpic}[width=0.24\linewidth]{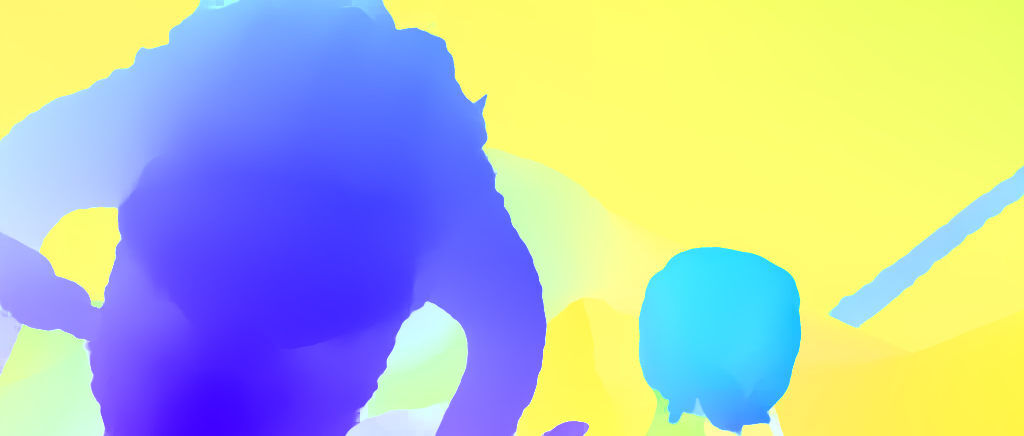}
        \put(34,-6.0){\scriptsize{(a) Frame 34}}
    \end{overpic}%
    \begin{overpic}[width=0.24\linewidth]{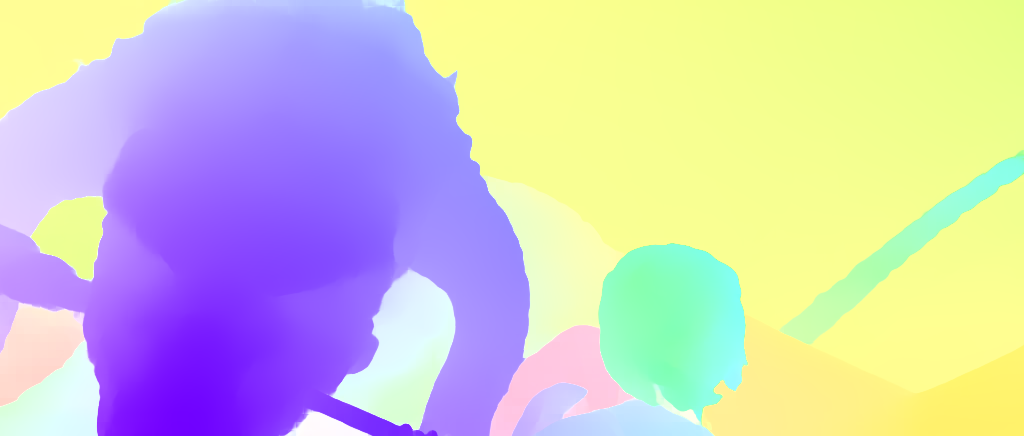}
        \put(34,-6.0){\scriptsize{(b) Frame 35}}
    \end{overpic}%
    \begin{overpic}[width=0.24\linewidth]{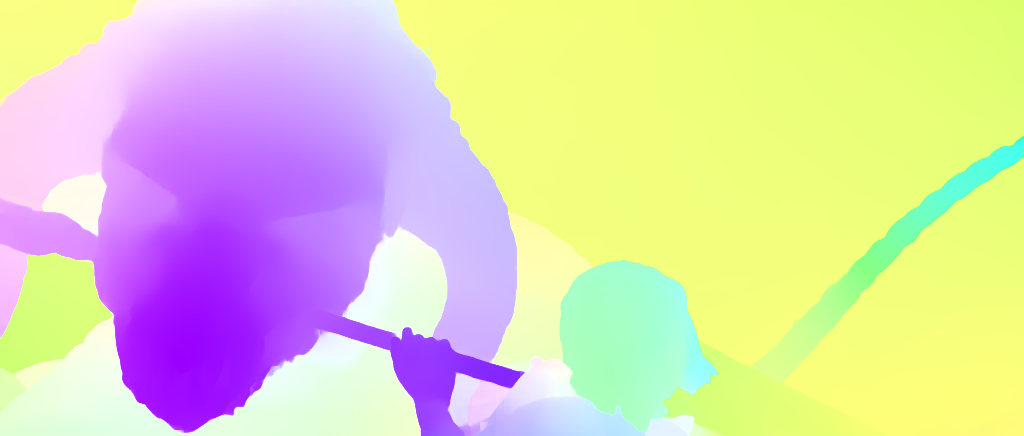}
        \put(34,-6.0){\scriptsize{(c) Frame 36}}
    \end{overpic}%
    \begin{overpic}[width=0.24\linewidth]{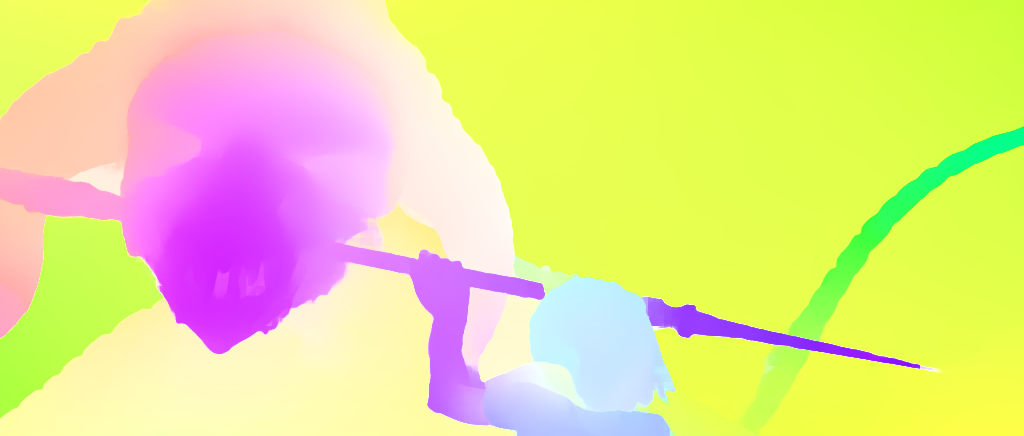}
        \put(34,-6.0){\scriptsize{(d) Frame 37}}
    \end{overpic}%
    \caption{
      Visualization on the Sintel test set, \texttt{cave\_3} sequence of the clean split.
    }
    \label{fig:demo-clean-cave-3}
    \vspace{-6pt}
\end{figure*}

\begin{figure*}[!ht]
    \centering
    \begin{overpic}[width=0.24\linewidth]{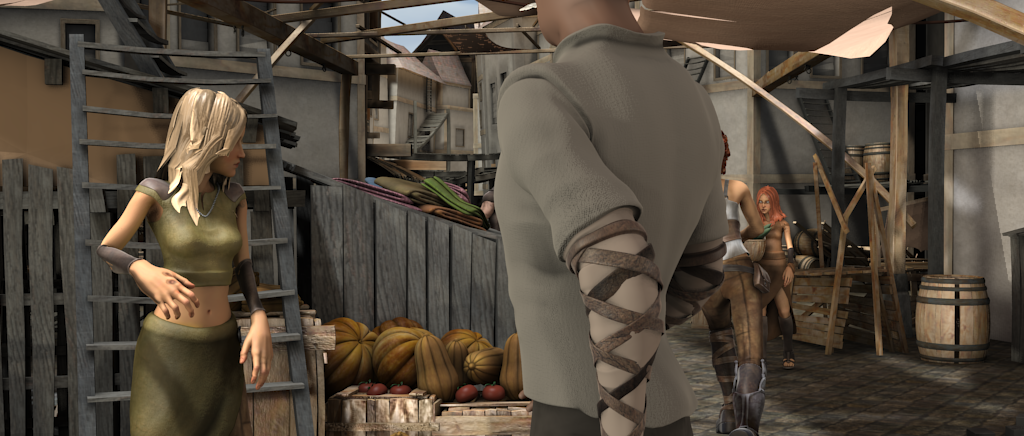}
    \end{overpic}%
    \begin{overpic}[width=0.24\linewidth]{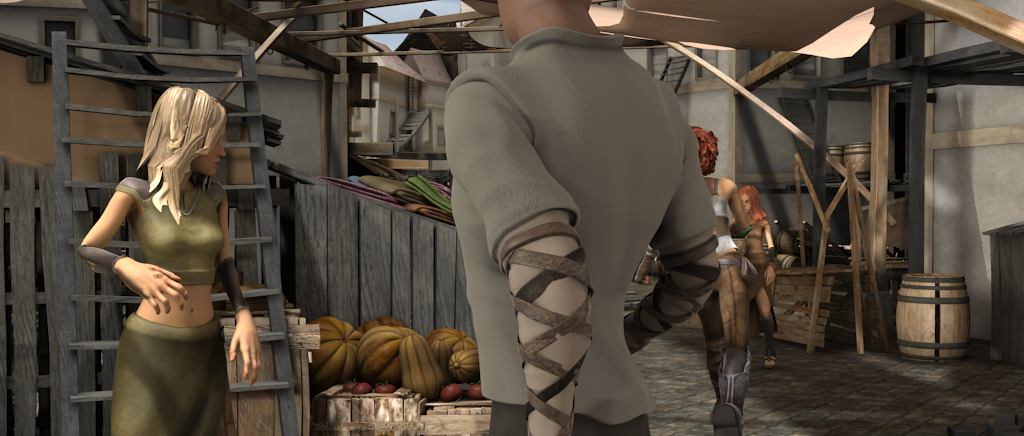}
    \end{overpic}%
    \begin{overpic}[width=0.24\linewidth]{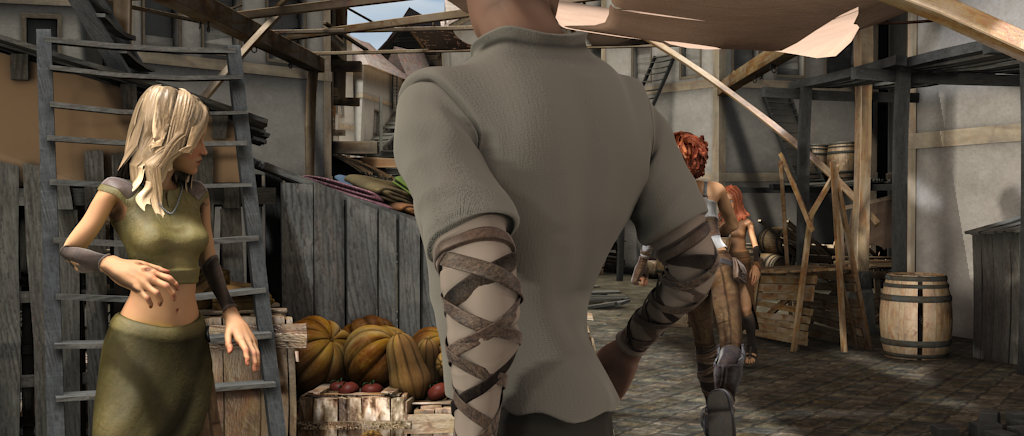}
    \end{overpic}%
    \begin{overpic}[width=0.24\linewidth]{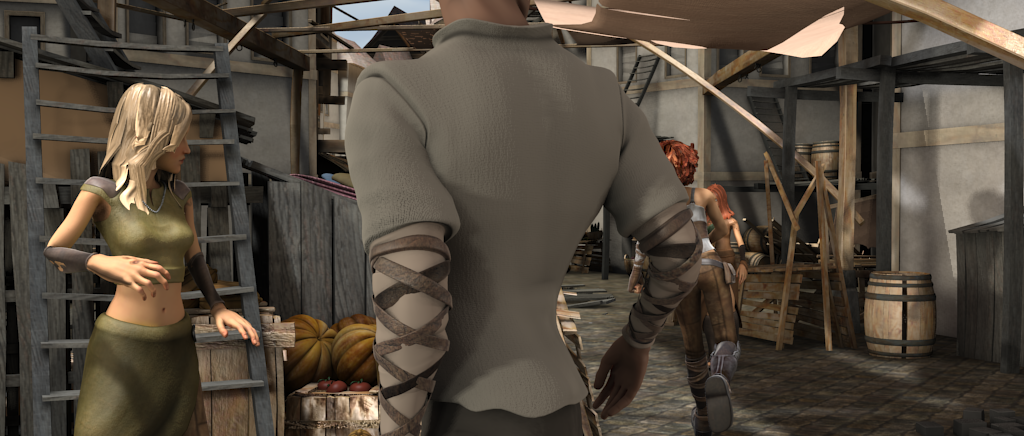}
    \end{overpic}%
    \\
    \begin{overpic}[width=0.24\linewidth]{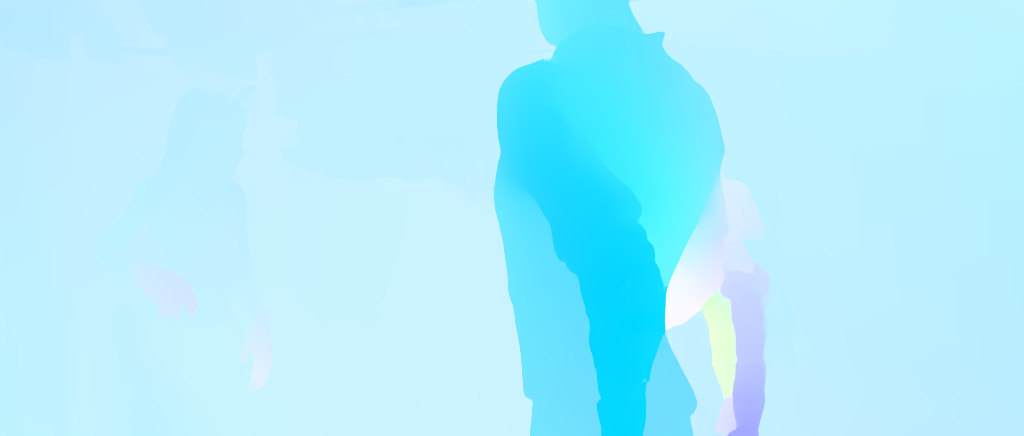}
        \put(34,-6.0){\scriptsize{(a) Frame 25}}
    \end{overpic}%
    \begin{overpic}[width=0.24\linewidth]{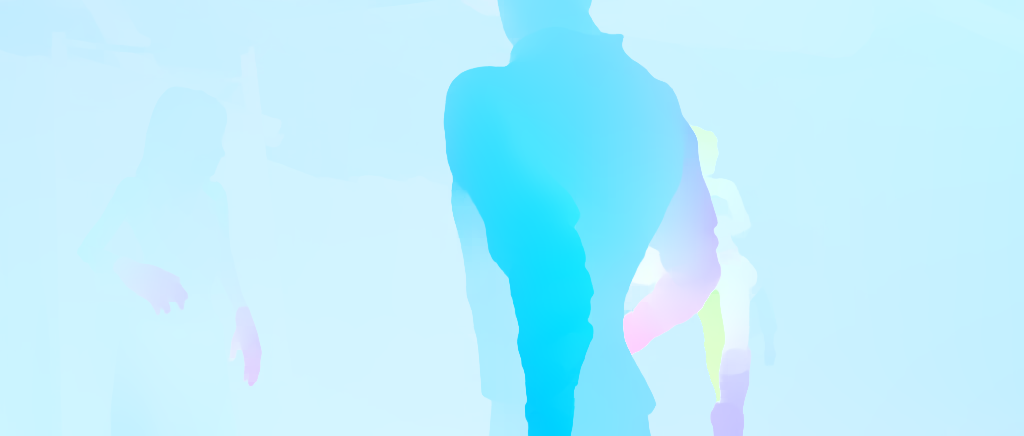}
        \put(34,-6.0){\scriptsize{(b) Frame 26}}
    \end{overpic}%
    \begin{overpic}[width=0.24\linewidth]{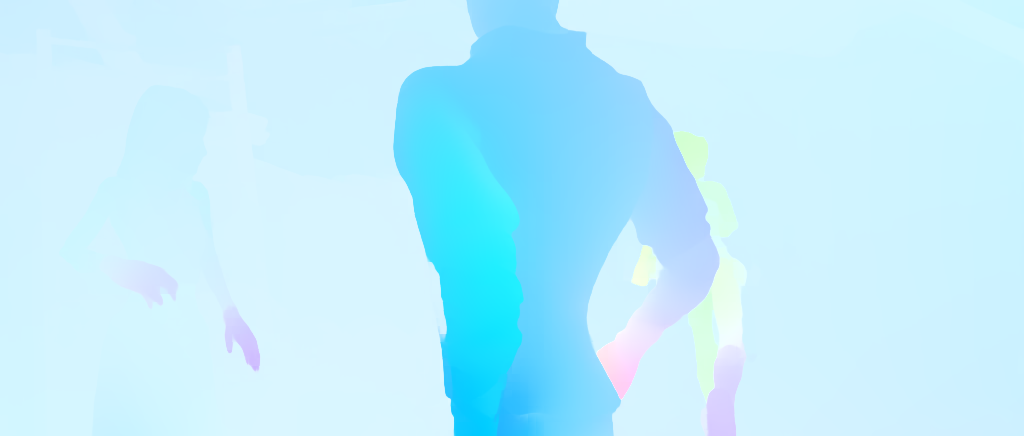}
        \put(34,-6.0){\scriptsize{(c) Frame 27}}
    \end{overpic}%
    \begin{overpic}[width=0.24\linewidth]{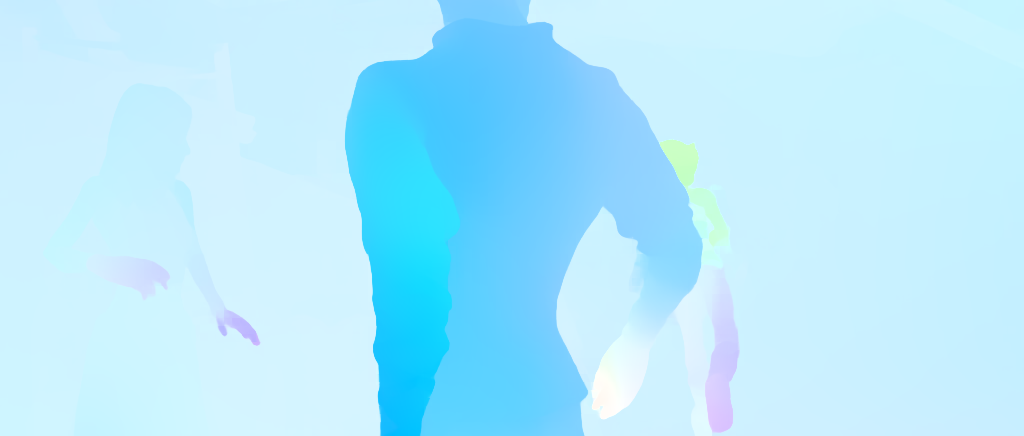}
        \put(34,-6.0){\scriptsize{(d) Frame 28}}
    \end{overpic}%
    \caption{
      Visualization on the Sintel test set, \texttt{market\_1} sequence of the clean split.
    }
    \label{fig:demo-clean-market-1}
    \vspace{-6pt}
\end{figure*}

\begin{figure*}[!ht]
    \centering
    \begin{overpic}[width=0.24\linewidth]{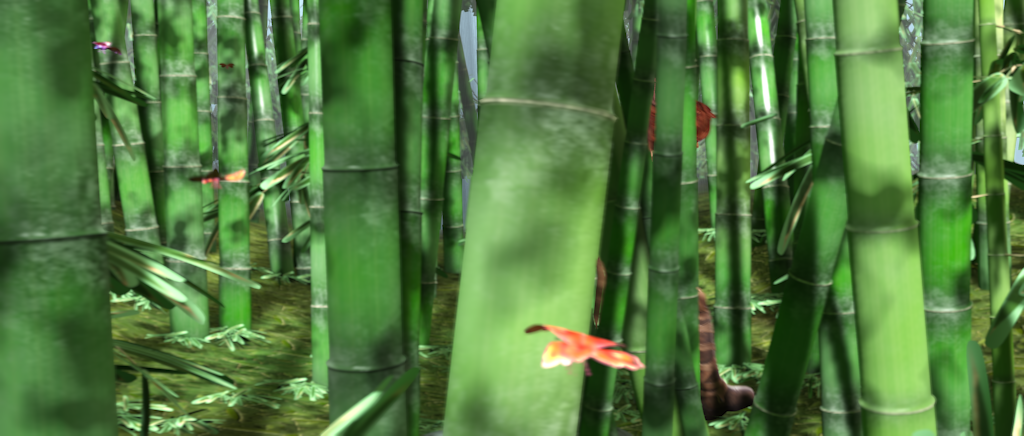}
    \end{overpic}%
    \begin{overpic}[width=0.24\linewidth]{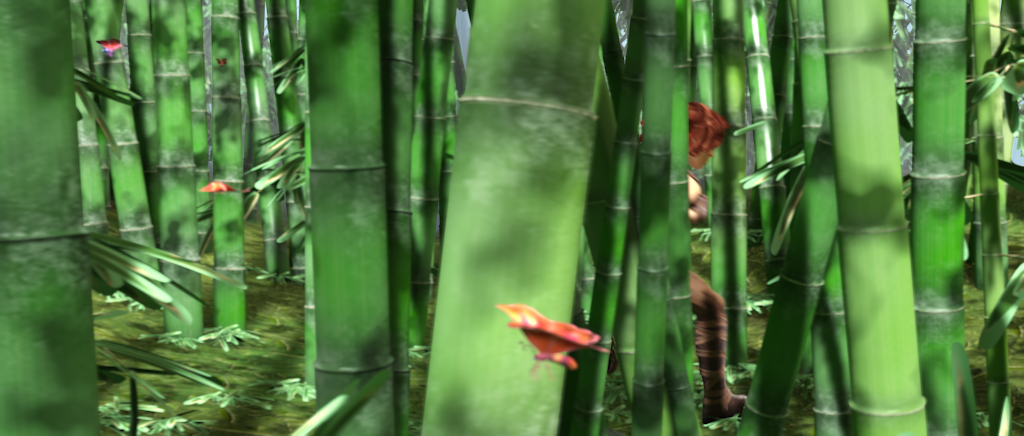}
    \end{overpic}%
    \begin{overpic}[width=0.24\linewidth]{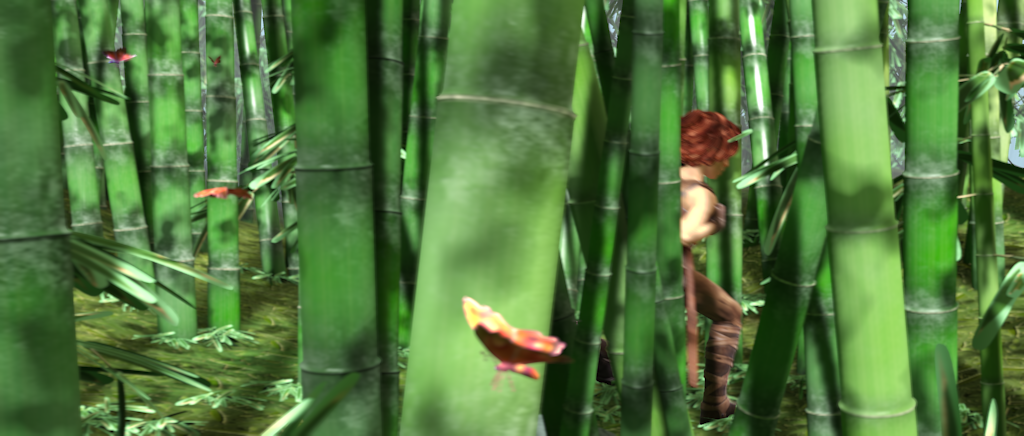}
    \end{overpic}%
    \begin{overpic}[width=0.24\linewidth]{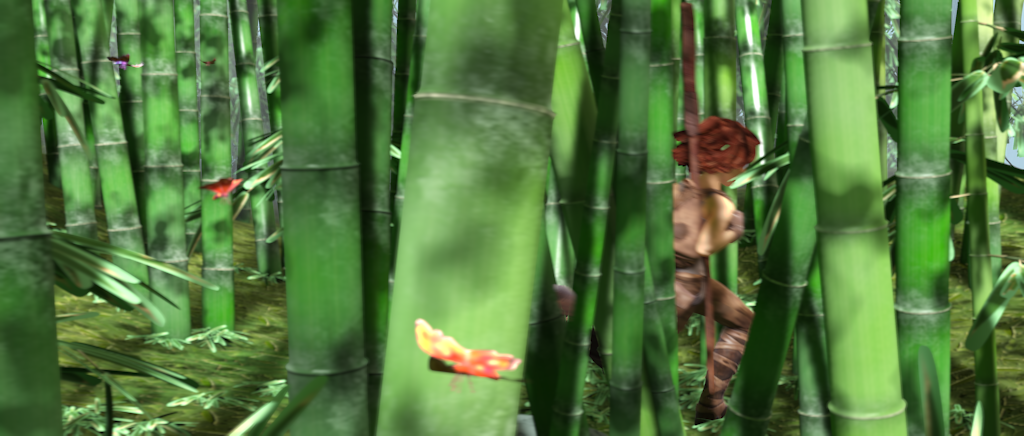}
    \end{overpic}%
    \\
    \begin{overpic}[width=0.24\linewidth]{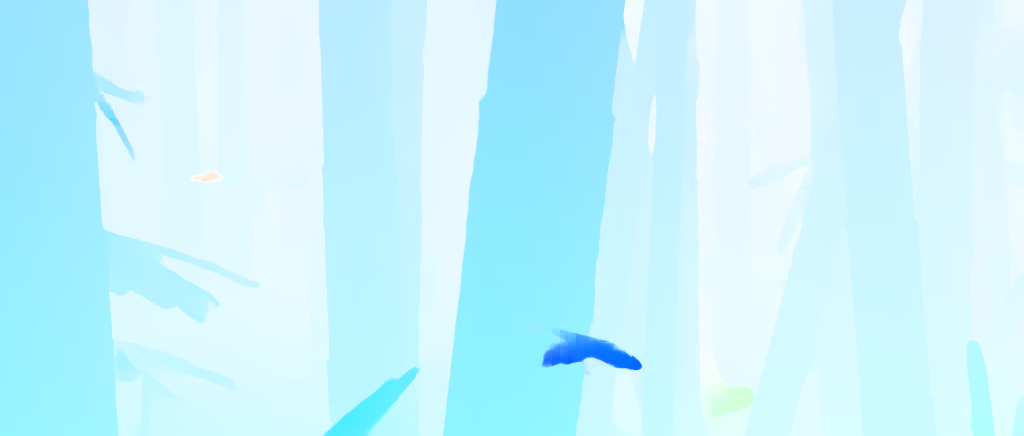}
        \put(34,-6.0){\scriptsize{(a) Frame 25}}
    \end{overpic}%
    \begin{overpic}[width=0.24\linewidth]{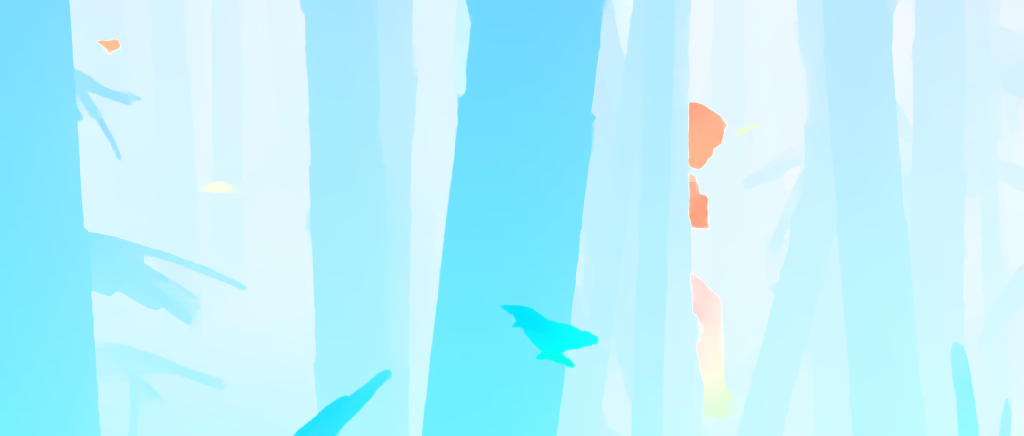}
        \put(34,-6.0){\scriptsize{(b) Frame 26}}
    \end{overpic}%
    \begin{overpic}[width=0.24\linewidth]{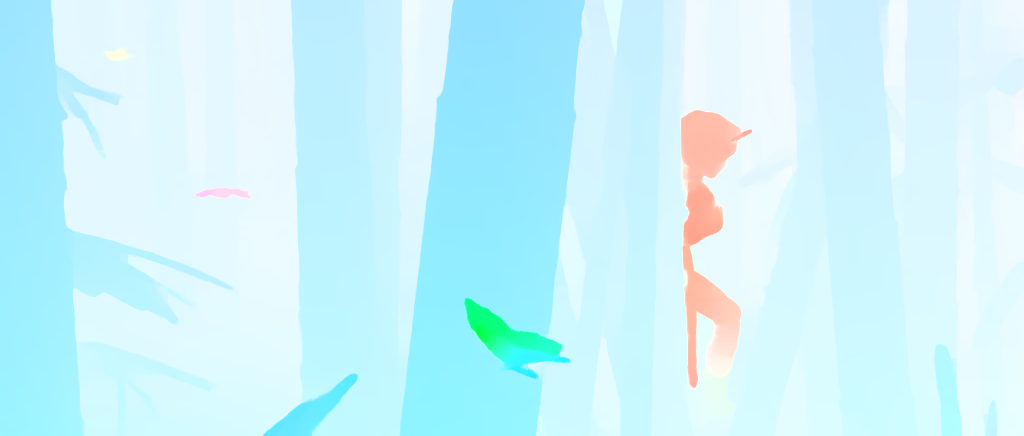}
        \put(34,-6.0){\scriptsize{(c) Frame 27}}
    \end{overpic}%
    \begin{overpic}[width=0.24\linewidth]{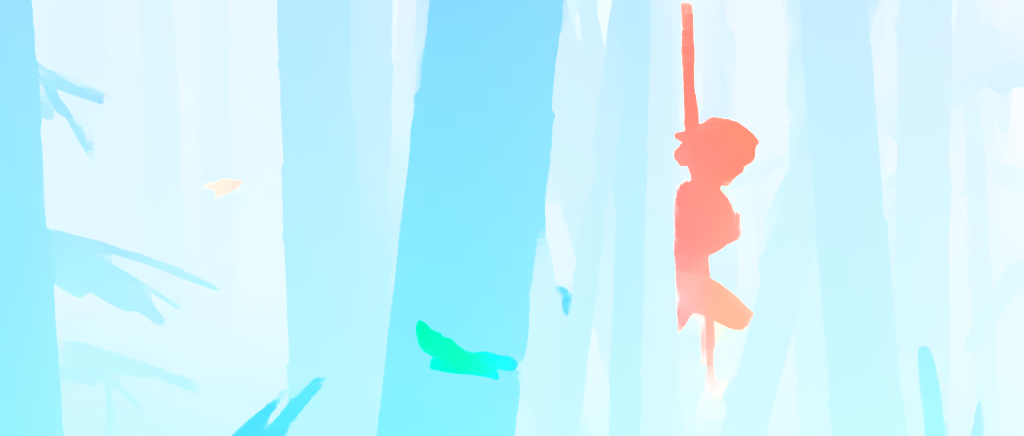}
        \put(34,-6.0){\scriptsize{(d) Frame 28}}
    \end{overpic}%
    \caption{
      Visualization on the Sintel test set, \texttt{bamboo\_3} sequence of the final split.
    }
    \label{fig:demo-final-bamboo-3}
    \vspace{-6pt}
\end{figure*}

\begin{figure*}[!ht]
    \centering
    \begin{overpic}[width=0.24\linewidth]{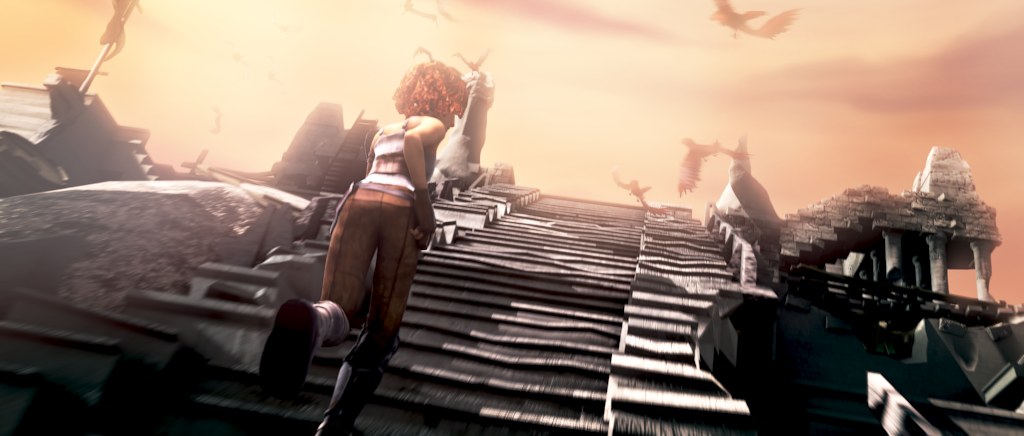}
    \end{overpic}%
    \begin{overpic}[width=0.24\linewidth]{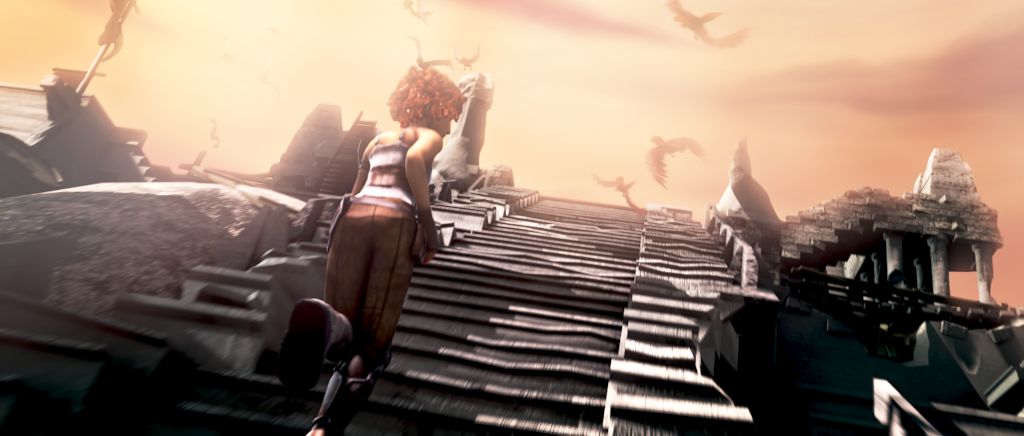}
    \end{overpic}%
    \begin{overpic}[width=0.24\linewidth]{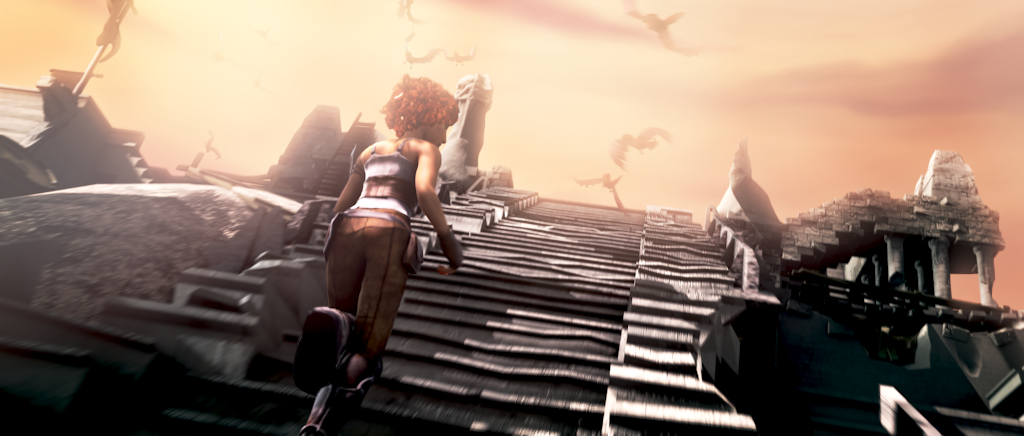}
    \end{overpic}%
    \begin{overpic}[width=0.24\linewidth]{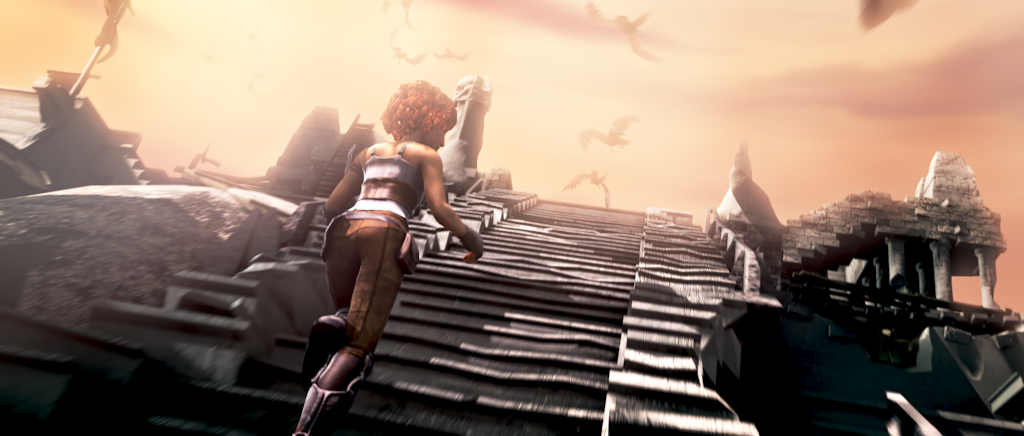}
    \end{overpic}%
    \\
    \begin{overpic}[width=0.24\linewidth]{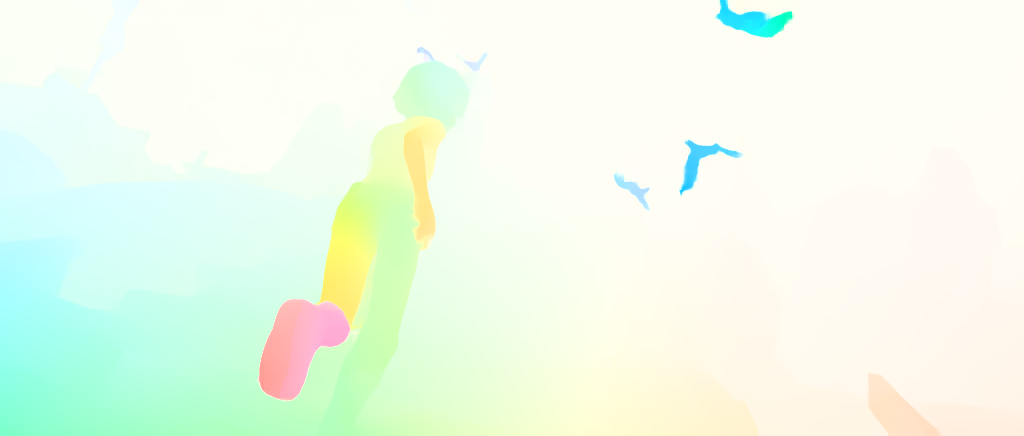}
        \put(34,-6.0){\scriptsize{(a) Frame 25}}
    \end{overpic}%
    \begin{overpic}[width=0.24\linewidth]{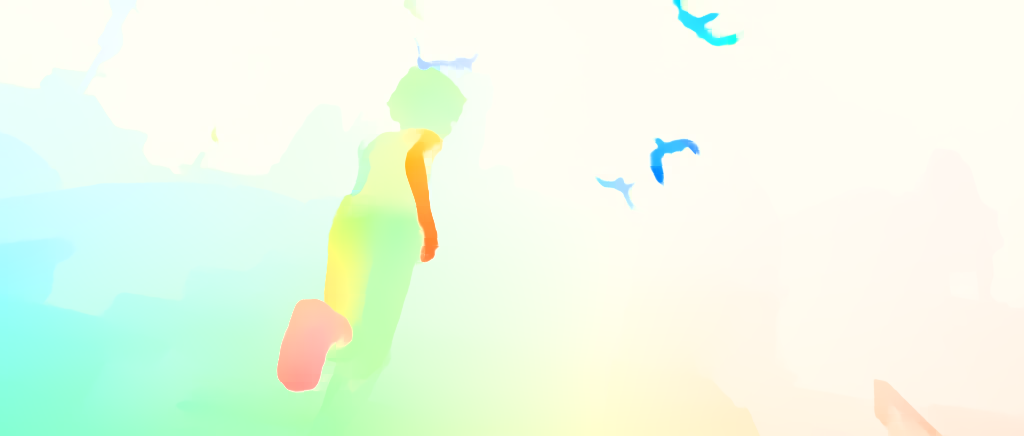}
        \put(34,-6.0){\scriptsize{(b) Frame 26}}
    \end{overpic}%
    \begin{overpic}[width=0.24\linewidth]{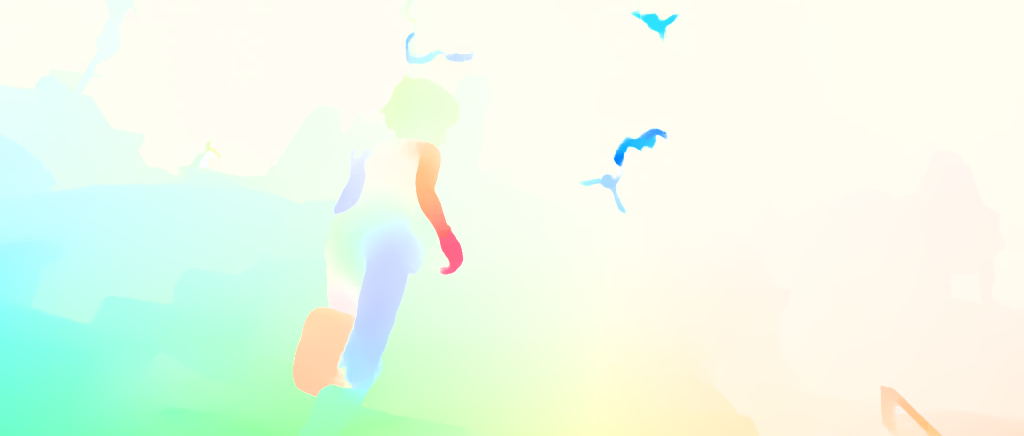}
        \put(34,-6.0){\scriptsize{(c) Frame 27}}
    \end{overpic}%
    \begin{overpic}[width=0.24\linewidth]{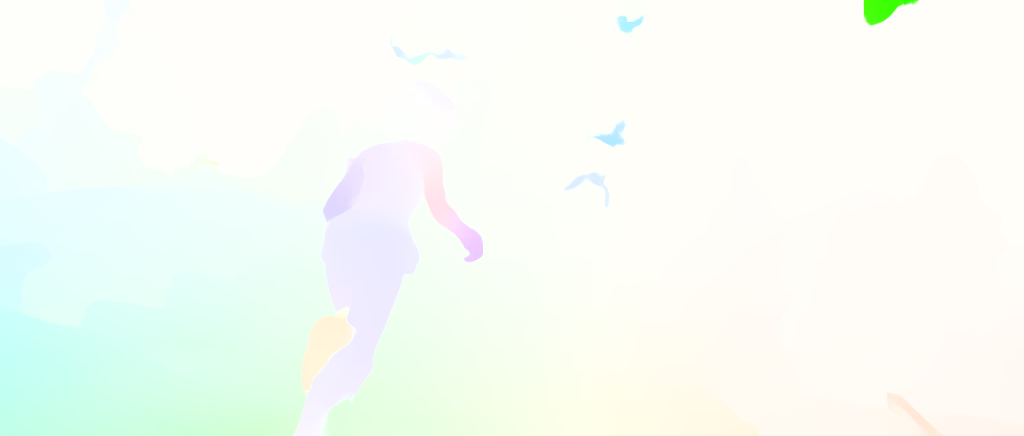}
        \put(34,-6.0){\scriptsize{(d) Frame 28}}
    \end{overpic}%
    \caption{
      Visualization on the Sintel test set, \texttt{temple\_1} sequence of the final split.
    }
    \label{fig:demo-final-temple-1}
    \vspace{-6pt}
\end{figure*}

\begin{figure*}[!ht]
    \centering
    \begin{overpic}[width=0.9\linewidth]{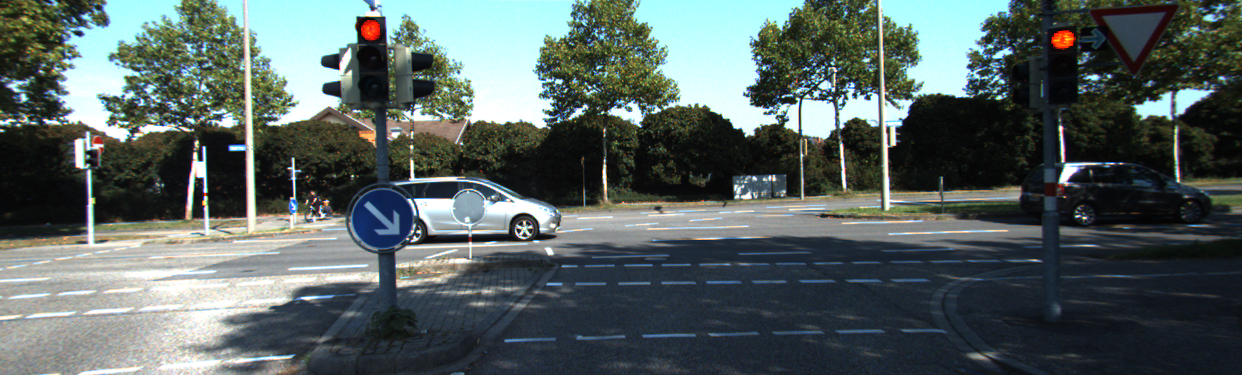}
        \put(46.5, -3.75){\scriptsize{Input frame}}
    \end{overpic}%
    \\ 
    \vspace{1cm}
    \begin{overpic}[width=0.45\linewidth]{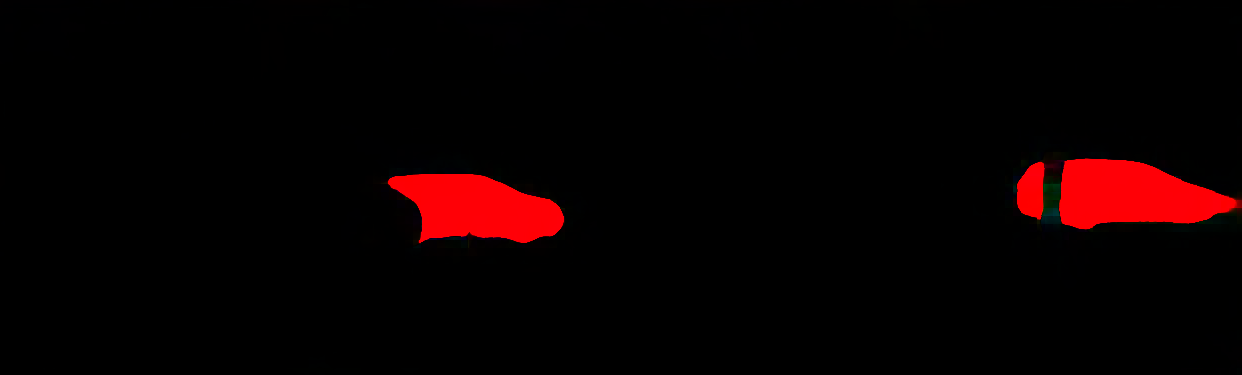}
        \put(42, -7.5){\scriptsize{RAFT prediction}}
    \end{overpic}%
    \begin{overpic}[width=0.45\linewidth]{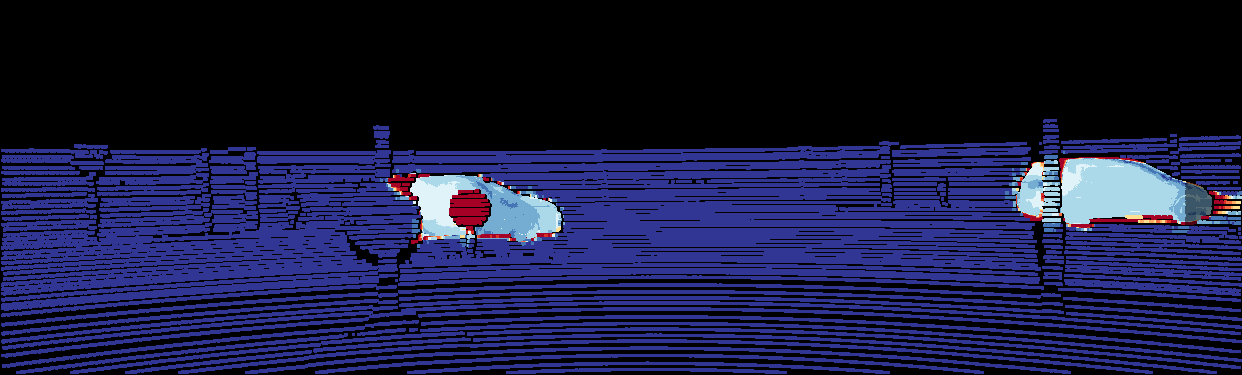}
        \put(43, -7.5){\scriptsize{RAFT error}}
    \end{overpic}%
    \\ 
    \vspace{1cm}
    \begin{overpic}[width=0.45\linewidth]{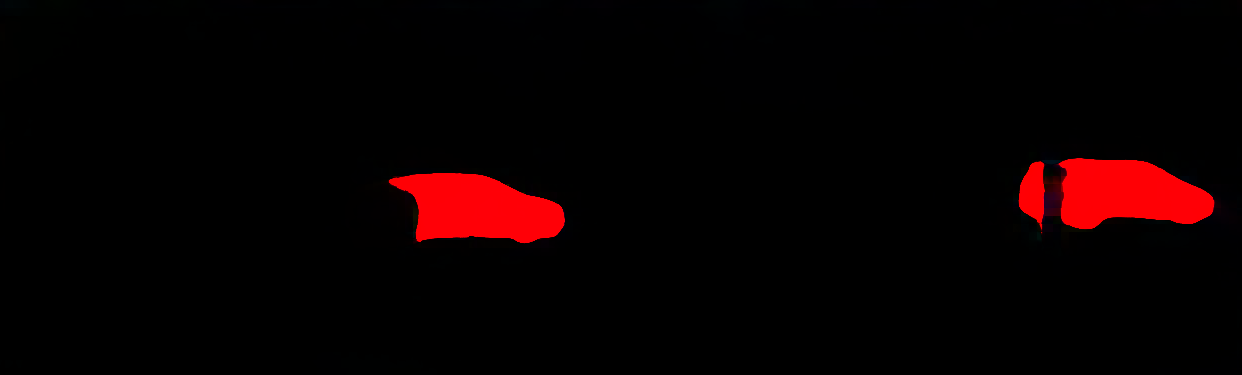}
        \put(41.8, -7.5){\scriptsize{GMA prediction}}
    \end{overpic}%
    \begin{overpic}[width=0.45\linewidth]{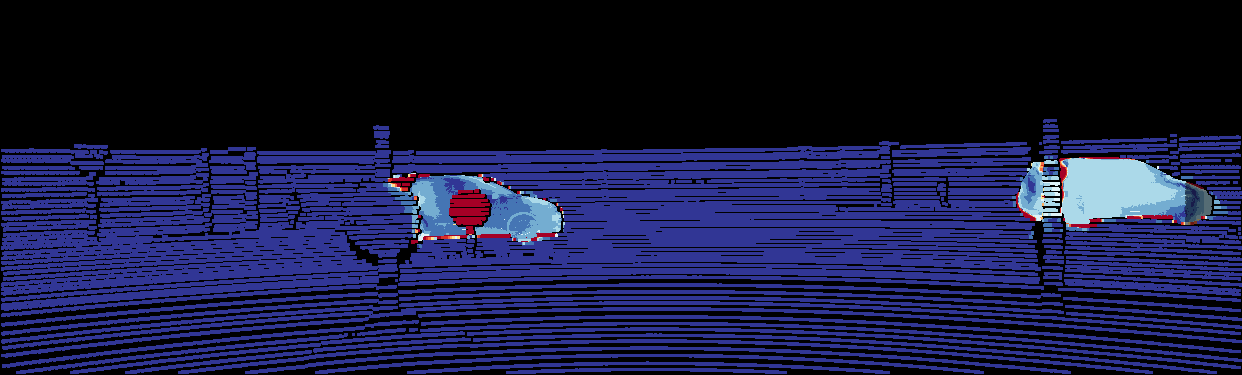}
        \put(42.8, -7.5){\scriptsize{GMA error}}
    \end{overpic}%
    \\ 
    \vspace{1cm}
    \begin{overpic}[width=0.45\linewidth]{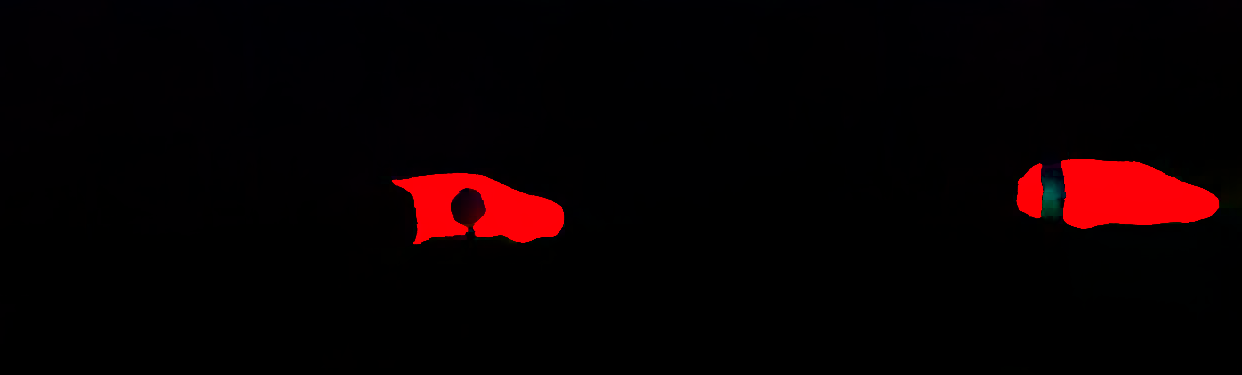}
        \put(38.7, -7.5){\scriptsize{DEQ Flow prediction}}
    \end{overpic}%
    \begin{overpic}[width=0.45\linewidth]{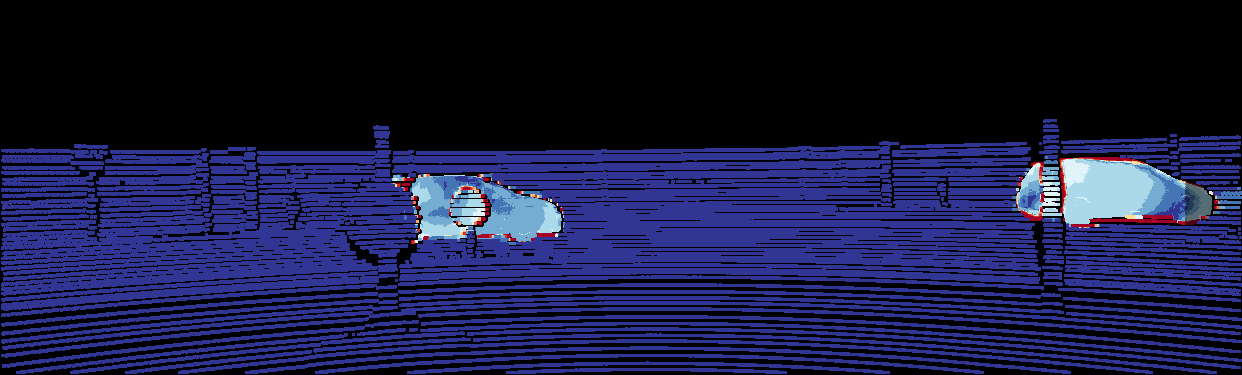}
        \put(39.4, -7.5){\scriptsize{DEQ Flow error}}
    \end{overpic}%
    \\ 
    \vspace{0.7cm}
    \caption{
      Visualization on the KITTI test set.
    }
    \label{fig:demo-kitti}
\end{figure*}

We visualize the flow estimation by the DEQ flow model in \cref{fig:demo-clean-ambush-1}, \cref{fig:demo-clean-cave-3}, \cref{fig:demo-clean-market-1}, \cref{fig:demo-final-bamboo-3}, \cref{fig:demo-final-temple-1}, and \cref{fig:demo-kitti} using consecutive frames of the MPI Sintel~\cite{sintel} test set and KITTI~\cite{kitti} test set. Flow estimation errors are downloaded from the leaderboard.


\end{document}